\documentclass{article} 
\usepackage{collas2024_conference,times}
\usepackage{easyReview}
\usepackage{sepfootnotes}
\usepackage{url}
\usepackage{multirow}
\usepackage{booktabs}
\usepackage{algorithm}
\usepackage{algpseudocode}
\usepackage{subcaption}
\usepackage[labelfont=bf]{caption}
\usepackage{minitoc}
\usepackage{adjustbox}
\usepackage{listings}
\usepackage{xcolor}

\definecolor{codegreen}{rgb}{0,0.6,0}
\definecolor{codegray}{rgb}{0.5,0.5,0.5}
\definecolor{codepurple}{rgb}{0.58,0,0.82}
\definecolor{backcolour}{gray}{0.95}
\definecolor{framecolour}{rgb}{0.8,0.8,0.8}

\lstdefinestyle{mystyle}{
    backgroundcolor=\color{backcolour},   
    commentstyle=\color{codegreen},
    keywordstyle=\color{magenta},
    numberstyle=\tiny\color{codegray},
    stringstyle=\color{codepurple},
    basicstyle=\ttfamily\footnotesize,
    breakatwhitespace=false,         
    breaklines=true,                 
    captionpos=b,                    
    keepspaces=true,                 
    numbers=left,                    
    numbersep=5pt,                  
    showspaces=false,                
    showstringspaces=false,
    breakautoindent=false,
    framerule=1pt,
    showtabs=false,                  
    tabsize=2,
    rulecolor=\color{framecolour},
    framextopmargin=5pt,
    framexbottommargin=5pt,
    framesep=5pt,
    resetmargins=true
}
\collasfinalcopy

\usepackage{amsmath,amsfonts,bm}

\def\eqref#1{equation~\ref{#1}}

\def\1{\bm{1}}

\DeclareMathAlphabet{\mathsfit}{\encodingdefault}{\sfdefault}{m}{sl}
\SetMathAlphabet{\mathsfit}{bold}{\encodingdefault}{\sfdefault}{bx}{n}

\usepackage{hyperref}
\hypersetup{
    colorlinks=true,
    linkcolor=red,
    filecolor=magenta,
    urlcolor=blue,
    citecolor=purple,
    pdftitle={Overleaf Example},
    pdfpagemode=FullScreen,
}

\title{\centering SymbolicAI: A framework for logic-based approaches \\ combining generative models and solvers}

\author{%
  ~~~~~~~~~~~~~~~~Marius–Constantin Dinu\textsuperscript{*~\textdagger~\textdaggerdbl}~~~~~~~~~Claudiu Leoveanu–Condrei\textsuperscript{\textdagger~\textbardbl}~~~~~~~~~Markus Holzleitner\textsuperscript{\textdaggerdbl}\\~~~~~~~~~~~~~~~~~~~~~~~~~~~~~~~~~~~~~~~~~~~~~~~~~~~~~~~\textbf{Werner Zellinger}\textsuperscript{\textdaggerdbl~\textsection}~~~~~~~~~~\textbf{Sepp Hochreiter}\textsuperscript{\textdaggerdbl} \\~~~~~~~~~~~~~~~~~~~~~~~ExtensityAI\textsuperscript{\textdagger}~~~~~~~~~Johannes Kepler University\textsuperscript{\textdaggerdbl}~~~~~~~~~RICAM\textsuperscript{\textsection}~~~~~~~~~Amazon Devices\textsuperscript{\textbardbl}
}

\newcommand{\ttt}{\texttt}

\begin{document}

\maketitle

\begin{abstract}
We introduce \emph{SymbolicAI}, a versatile and modular framework employing a logic-based approach to concept learning and flow management in generative processes.
SymbolicAI enables the seamless integration of generative models with a diverse range of solvers by treating large language models (LLMs) as semantic parsers that execute tasks based on both natural and formal language instructions, thus bridging the gap between symbolic reasoning and generative AI. 
We leverage probabilistic programming principles to tackle complex tasks, and utilize differentiable and classical programming paradigms with their respective strengths. 
The framework introduces a set of polymorphic, compositional, and self-referential operations for multi-modal data that connects multi-step generative processes and aligns their outputs with user objectives in complex workflows.
As a result, we can transition between the capabilities of various foundation models with in-context learning capabilities and specialized, fine-tuned models or solvers proficient in addressing specific problems. 
Through these operations based on in-context learning our framework enables  the creation and evaluation of explainable computational graphs.
Finally, we introduce a quality measure and its empirical score for evaluating these computational graphs, and propose a benchmark that compares various state-of-the-art LLMs across a set of complex workflows. 
We refer to the empirical score as the "Vector Embedding for Relational Trajectory Evaluation through Cross-similarity", or \emph{VERTEX}~score for short. 
The \href{https://github.com/ExtensityAI/symbolicai}{framework codebase}\footnote{\ SymbolicAI framework: \url{https://github.com/ExtensityAI/symbolicai}} and \href{https://github.com/ExtensityAI/benchmark}{benchmark}\footnote{\ Evaluation benchmark: \url{https://github.com/ExtensityAI/benchmark}\\\text{~~~~~*~}Correspondence to: \href{mailto:dinu@ml.jku.at}{\texttt{dinu@ml.jku.at}}, \texttt{\{\href{mailto:marius@extensity.ai}{marius}, \href{mailto:leo@extensity.ai}{leo}\}@extensity.ai}\\\text{~~~~~~\textbardbl~} Work done outside of Amazon.} are linked below.

\end{abstract}

\begin{figure}[h!]
    \centering
    \includegraphics[width=1.0\linewidth]{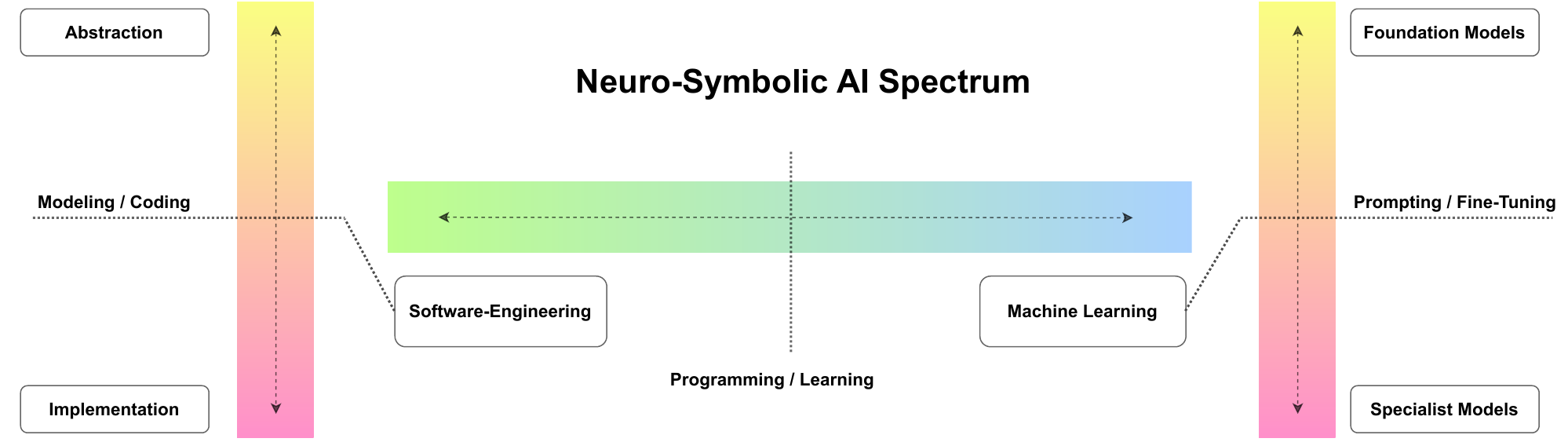}
    \caption{Our neuro-symbolic framework enables a seamless transition between symbolic and differentiable programming, each with distinct dynamics and strengths. Differentiable programming provides access to foundational and specialist models. Classical programming, on the other hand, shifts between abstraction and implementation, focusing on high-level concepts before delving into the details of implementation.}
    \label{fig:nsai_spectrum}
\end{figure}

 \pagebreak
\section{Introduction}
\label{sec:introduction}

The recent surge in generative AI, particularly involving large language models (LLMs), has demonstrated their wide-ranging applicability across various domains \citep{Badita:22, Degrave:22}.
These models have enhanced the functionality of tools for search-based interactions \citep{You.com:22, Chatsonic:22, Bing:23}, program synthesis \citep{Jain:21, Parades:23, Key:23}, chat-based interactions \citep{Replika:16, OpenAIChatGPT:22, Google:23}, and many more.
Moreover, language-based approaches have facilitated connections between different modalities, enabling text-to-image \citep{Ramesh:21, Saharia:22}, text-to-video \citep{Singer:22}, text-to-3D \citep{Poole:22}, text-to-audio \citep{Oord:16, Wang:17}, and text-to-code \citep{Wang:21, Lu:21, Li:22} transformations, to name a few.
Consequently, by training on vast quantities of unlabelled textual data, LLMs have been shown to not only store factual knowledge \citep{Petroni:2019,Kassner:20} and approximate users’ intentions to some extent \citep{Andreas:22}, but also to unlock deep specialist capabilities through innovative prompting techniques \citep{Nori:23}.

Despite their versatility, current LLMs face challenges such as fallacious reasoning and the generation of erroneous content, commonly referred to as hallucinations \citep{Jones:22}. 
These limitations highlight the importance of integrating complementary symbolic methods to validate and guide the generative processes of LLMs, ensuring more accurate and reliable outputs.
In parallel, efforts have focused on developing tool-based approaches \citep{Schick:23} or template frameworks \citep{Chase:23} to extend LLMs' capabilities and enable a broader spectrum of applications.
However, these efforts only partially capture the potential inherent in leveraging LLMs as \emph{semantic parsers}. 
In contrast to parsers for structured languages a semantic parser is able to break down unstructured human language into semantically meaningful components and transform those into a structured form. 
While traditionally semantic parsing has been a role filled by specialized algorithms and models, we posit that LLMs, through their training on diverse linguistic data, have developed the ability to perform semantic parsing as part of their broader natural language processing capabilities.
In turn, we identify LLMs as a central component in creating sophisticated neuro-symbolic (NeSy) AI systems.
These systems integrate symbolic and sub-symbolic concepts and utilize the capabilities of semantic parsing to develop symbolic expressions that enable new probabilistic programming paradigms.

We introduce \emph{SymbolicAI}, a compositional NeSy framework able to represent and manipulate multi-modal and self-referential structures  \citep{Schmidhuber:07, Fernando:23}.
SymbolicAI augments the generative process of LLMs with in-context learning operations, realized through functional primitives, and enables the creation of versatile applications through in-context learning \citep{Wei:22}.
These operations enable logic-based components that guide the generative process and enable a modular NeSy system, including a wide range of existing solvers, formal language engines for mathematical expression evaluation, theorem provers, knowledge bases, and search engines for information retrieval.
SymbolicAI exposes these solvers as building blocks for constructing compositional functions as computational graphs, making it possible to bridge classical and differentiable programming paradigms with the aim to create \emph{domain-invariant problem solvers}.
In designing the architecture of SymbolicAI, we drew inspiration from a body of evidence that suggests the human brain possesses a selective language processing module \citep{Macsweeney:02,Fedorenko:10,Menenti:11,Regev:13,Terri:16,Deniz:19,Hu:21}, prior research on cognitive architectures \citep{Newell:56, Newell:57, Newell:72, Newell:90, Laird:22}, and the significance of language on the structure of semantic maps in the human brain \citep{Huth:16}. 
We consider language as a central processing module, distinct from other cognitive processes such as reasoning or memory \citep{Paischer:22, Paischer:23}.
We hypothesize that such a central processing module based in language is a core component of broad AI systems (see Appendix Section~\ref{sec:broad_ai}) and enables the development of fully autonomous AI systems for decision-making.

A significant challenge encountered in the development of our framework pertained to the evaluation of LLMs when used as semantic parsers in a NeSy workflow. 
Current evaluation of generated content relies on metrics for single-step generative processes, such as the BLEU score \citep{Papineni:02}.
These metrics are not suitable for evaluating multi-step generative processes.
BLEU has limitations, as it measures $n$-gram-based overlap of generated output with a reference that does not consider the semantic meaning.
As a result, BLEU fails to capture semantic equivalence, especially in more complex tasks.
More recent metrics such as CIDEr~\citep{Vedantam:14} or SPICE~\citep{Anderson:16} are also not suitable for our problem, either because they are built upon BLEU (in case of CIDEr) or designed with inductive biases specifically for image captioning.

Therefore, alongside our framework we introduce a quality measure (and its empirical score) for multi-step generative processes based on semantic meaning.
We refer to our score as "Vector Embedding for Relational Trajectory Evaluation through Cross-similarity", or VERTEX score for short.
Our VERTEX score uses embeddings to compare node distributions within a computational graph.
It measures the semantic meaning across the distributional path by computing at each node the cross-similarity between the generated embeddings and embeddings sampled from a reference distribution.
Furthermore, the VERTEX score is designed such that it can be used as a reward signal in a reinforcement learning setting \citep{Sutton:84}.
Finally, we propose a benchmark for evaluating complex workflows.
We define a set of basic evaluations, particularly associative predictions based on in-context learning, multi-modal bindings for tool utilization, and program synthesis for subroutine execution.
Furthermore, we introduce complex evaluations for logic-based components and hierarchical computational graphs.

In summary, the key contributions presented in this work are as follows:
\begin{itemize}
\item  We introduce SymbolicAI, a logic-based framework for concept learning and flow management in generative processes, enabling seamless integration with a wide range of foundation models and solvers.

\item We leverage LLMs as semantic parsers to enable the creation of complex computational graphs by combining symbolic expressions with probabilistic programming paradigms.

\item  We introduce a quality measure and its empirical score alongside a benchmark designed for multi-step generative processes for comparing LLMs across a wide range of complex tasks.

\end{itemize}

\section{Related Work}
\label{sec:related_work}

\paragraph{Symbolic Methods}
The field of symbolic AI has its foundations in the works of the Logic Theorist (LT) \citep{Newell:56} and the General Problem Solver (GPS) \citep{Newell:57}. 
These programs represented the first steps towards automated reasoning and problem-solving utilizing symbolic representations.
Despite their advancements, both faced challenges in dealing with the complexity of real-world problems, particularly due to the combinatorial nature of the solution space.
To address these limitations, the Soar \citep{Laird:87} cognitive architecture was developed, advancing the notion that intelligent behavior results from goal-oriented search through a problem space \citep{Newell:72, McCarthy:06}, with each step consisting of selecting and applying operators. 
Soar introduced components like reinforcement learning, impasses, sub-states, and chunking to enhance its problem-solving capabilities. 
It also demonstrated the importance of learning from experiences to adapt and improve performance over time.
However, \cite{Santoro:22} emphasizes the subjectivity of symbols and suggests that human-like symbolic fluency could develop in machines through learning algorithms immersed in socio-cultural contexts. 
This perspective, anchored in the notion that symbols are triadic and their meaning emerges from consensus, seeks to move away from traditional symbolic AI methodologies towards AI that adaptively learns meaning and behaviors from human-like experiences.
The goal is to cultivate machines that demonstrate symbolic behaviors across a spectrum of competencies, potentially mirroring the evolutionary and social learning processes observed in humans.
Lastly, symbolic AI struggles with real-world data's unpredictability and variability. 
These challenges have led to the employment of statistical learning methodologies, like deep learning \citep{Alom:18}, which are more adept at managing noise and uncertain information through vector-valued representations. 

\paragraph{Sub-Symbolic Methods}
The sub-symbolic framework, rooted in neural network paradigms, began with pioneering works such as the perceptron \citep{McCulloch:43}, with the first hardware implementation quickly following \citep{Rosenblatt:58}. 
The foundational notion of distributed processing \citep{Rumelhart:86} was later bolstered and further expanded by demonstrating that multilayer feedforward networks with a single hidden layer can serve as universal approximators for any Borel measurable function, given sufficient hidden units \citep{Hornik:89}.
Fast-forward, contemporary frameworks achieve a significant leap with the introduction of the Transformer architecture \citep{Vaswani:17}, which underpins most of today's LLMs. 
These LLMs demonstrate exceptional capabilities in in-context learning, a method popularized by the likes of GPT-3 \citep{Brown:20}, where models improve task performance through natural language instruction and examples provided directly in the input prompt.
While in-context learning bypasses the need for explicit retraining, it demands meticulous prompt design to steer models towards desired behaviors. 

\paragraph{Neuro-Symbolic Methods}
To overcome the limitations of each individual method, NeSy approaches meld the statistical inference strengths of deep neural architectures with the generalization and explainability of symbolic systems \citep{Garcez:15, Besold:17, Garcez:19, Garcez:20, Lamb:20, Hamilton:22, Yu:23}.
Some approaches focus on different strategies for integrating learning and reasoning processes \citep{Yu:23, Fang:24}. 
Firstly, \emph{learning for reasoning} methods treat the learning aspect as an accelerator for reasoning, in which deep neural networks are employed to reduce the search space for symbolic systems \citep{ Silver:16, Silver:17a, Silver:17b, Qu:19, Schrittwieser:20}. 
Secondly, \emph{reasoning for learning} views reasoning as a way to regularize learning, in which symbolic knowledge acts as a guiding constraint that oversees machine learning tasks \citep{Hu:16, Xu:18}. 
Thirdly, the \emph{learning-reasoning} category enables a symbiotic relationship between learning and reasoning. 
Here, both elements interact and share information to boost problem-solving capabilities \citep{Donadello:17, Manhaeve:18, Mao:19, Ellis:23}.
This synergy further extends when considering graph-based methods, which closely align with the objectives of our proposed framework. 
Research in this area, such as CycleGT \citep{Guo:20} and Paper2vec \citep{Ganguly:17} explores unsupervised techniques for bridging graph and text representations, GPTSwarm \citep{Zhuge:24} explores graph optimizers to refine node-level prompts and edge optimization. 
Subsequently, graph embeddings, when utilized within symbolic frameworks, can enhance knowledge graph reasoning tasks \citep{Zhang:21}, or more generally, provide the bedrock for learning domain-invariant representations \citep{Park:23}.

Lastly, building upon the insights from \cite{Sun:22}, the integration of NeSy techniques in scientific workflows promises significant acceleration in scientific discovery. 
While previous work has effectively identified opportunities and challenges, we have taken a more ambitious approach by developing a comprehensive framework from the ground up to facilitate a wide range of NeSy integrations.

\paragraph{Large Language Models}
In part, instruction-based fine-tuning of LLMs through reinforcement learning from human feedback \citep{Ouyang:22, Li:23} or direct preference optimization \citep{Rafailov:23} has shown promising results dealing with value misalignment issues \citep{Knox:08, Macglashan:17, Christiano:17, Ibarz:18, Goyal:22}, unlocking new possibilities for chain of thoughts \citep{Wei:22CoT}, tree of thoughts \citep{Yao:23}, and graph of thoughts interactions \citep{Besta:23}.
However, recent research also highlights the limitations of LLMs in functional linguistic competence despite their proficiency in formal linguistic competence \citep{Mahowald:23}.
Whereas formal linguistic competence encompasses the ability to understand and generate language, functional linguistic competence pertains to the application of language in real-world contexts, such as conveying sensory input or recalling information from memory. 
Examples of functional linguistic competence include implicatures \citep{Ruis:22} and contextual language comprehension beyond the statistical manifestation of data distributions \citep{Bransford:72, Mikolov:13}. 
Consequently, operating LLMs through a purely inference-based approach confines their capabilities within their provided context window, severely limiting their horizon. 
This results in deficiencies for situational modeling, non-adaptability through contextual changes, and short-term problem-solving, amongst other capabilities.
However, simply increasing the context length may not yield greater capabilities, as demonstrated by the observed U-shaped performance curve \citep{Liu:23} where LLMs excel when utilizing information at the beginning or end of the input context, but struggle with information located in the middle, especially as context increases.
These challenges are actively being researched, with novel approaches such as Hyena \citep{Poli:23}, RWKV \citep{Bo:21}, GateLoop \citep{Katsch:23}, Mamba \citep{Gu:23} and xLSTM \citep{Beck:24} surfacing.  
Meanwhile, the re-emergence of interest in retrieval-augmented generative approaches \citep{HuayangLi:22} offers an alternative by circumventing the autoregressive nature of the widely-utilized Transformer architecture \citep{Vaswani:17}, enabling context enrichment with lateral information.

\paragraph{In-Context Learning}
\label{sec:icl}
Recently, several in-context learning methodologies evolved to enable tool usage through LLMs \citep{Schick:23}, or refine the generative outcome of LLMs \citep{Yang:23}. 
This includes chain-of-thought (CoT) prompting, a method that conditions the model to reveal its step-by-step reasoning process \citep{Wei:22CoT, Singhal:23}. 
CoT prompting breaks down complex tasks into simpler, sequential steps, and helps with interpreting LLM's output. 
Self-generated CoT, where models are encouraged to generate their own reasoning chains based on training examples, surpasses even expertly crafted CoT \citep{Fernando:23}.
This observation echoes other reports that GPT-4 has an emergent self-improving capability through introspection, such as self-verification \citep{Weng:23} or self-consistency \citep{Wang:23consistency}.
Tree of Thoughts (ToT) enables LLMs to solve complex problems by exploring multiple reasoning paths through a search tree of coherent text units, demonstrating significant problem-solving enhancements in tasks requiring strategic planning and search \citep{Yao:23}.
Ensemble techniques further enhance the robustness and accuracy of model predictions by combining several strategies to establish a consensus \citep{Nori:23}.

\section{Problem Definition}

Conventional approaches employing foundation models, such as LLMs, are predominantly confined to single-step or few-step executions and primarily reliant on hand-crafted prompt instructions, often referred to as in-context learning.
This restricted scope limits the utilization of different modalities, lacks verification, and exhibits limited tool proficiency.
We posit that the use of NeSy engines as core computation units, realized through logic-based methodologies coupled with sub-symbolic foundation models, offers a more general, robust, and verifiable perspective.
This approach has several advantages. 
Firstly, it enables the integration of pre-existing solutions (e.g. various classical algorithms), offloading computational complexity and bridging different modalities. 
Secondly, it allows sub-symbolic components to focus on decision-making (e.g. selecting the respective tool based on in-context classification). 
Thirdly, it provides an \emph{interpretable language-based control layer} for explainable, autonomous systems.
In the following section, we elaborate on the key design principles underlying SymbolicAI and how we guide the generative processes of NeSy engines. For further technical details, see Appendix Section~\ref{sup:framework}.

\section{Design Principles}\label{sec:symandexpr}

\paragraph{Symbols and Expressions} As posited by \cite{NewellSimon:76}, symbols are elemental carriers of meaning within a computational context\footnote{\ Our framework's name is derived from the foundational work of Newell and Simon.}. 
These symbols define physical patterns capable of composing complex structures, and are central to the design and interpretation of logic and knowledge representations \citep{Augusto:22}.
We define a symbol as the set $\mathcal{S} = \bigcup_{n\geq0}\mathbb{L}^n$ formed by concatenating characters from a finite character set $\mathbb{L}$, i.e. the vocabulary in an LLM setting, and with $n$ representing the sequence length of the string.
Thus, let the set of all possible symbols be defined as $\Sigma$ and that $\mathcal{S} \in \Sigma$.
We further introduce an operation $\bigoplus$ that enables us to create expressions on any number of symbols from $\Sigma$, and when evaluated returns a new symbol in $\Sigma$.
For any subset $\{ \mathcal{S}_1, \mathcal{S}_2, \ldots, \mathcal{S}_m \} \subseteq \Sigma$, an expression is defined as $\omega : \bigoplus_{i=1}^{m} \mathcal{S}_i \rightarrow \mathcal{S}'$ from the set of all possible expressions $\omega \in \Omega$, where $\mathcal{S}' \in \Sigma$, and $\bigoplus$ represents the placeholder operation of combining and transforming the symbols according to specific rules for $m$ number of symbols.
Such a specific rule for $\bigoplus$ can define an arithmetic expression $\bigoplus := +$ where two symbols are added, i.e. $\omega := \texttt{"1"} + \texttt{"two"}$ which results in a new symbol $\texttt{"3"}$ or $\texttt{"three"}$.
Thus, SymbolicAI is based on the concept that symbols, and the expressions they form, are reflections of the information inherent in a NeSy system, and serve as surrogate for the interaction between the NeSy system and the problem space. 
Moreover, we argue that \emph{real patterns}~\citep{Dennett:91}, recurring and identifiable structures that coherently and reliably emerge in the data beyond mere randomness or noise, can be effectively realized through symbols.

Furthermore, we utilize language as a tool for mapping complex concepts, leveraging its inherent semantics and abstractions to describe states and properties of a problem at hand.
These mappings are universal, e.g. they may be utilized to define scene descriptions, long-horizon planning, acoustic properties, emotional states, physical conditions, etc. 
Therefore, language serves as a comprehensive, yet abstract framework to encapsulate meanings, and refer to it as the \emph{convex hull of the knowledge of our society}.
Subsequently, it is common to attribute existing physical objects with abstract concepts, as exemplified by our natural tendency to link tangible objects to colors and emotions, such as blending the color "\emph{red}" with "\emph{heart}", "\emph{warm}", and "\emph{passion}". 
This approach also anchors our work in the field of formal language theory, as we require a structured method to construct mappings from the world to language. 
Consequently, we use formal language structures, such as grammars, to systematically define our language-centric approach to problem-solving and the associated translation of real-world complexities into linguistic terms.

\paragraph{Formal Languages}\label{formal_languages}
In formal language theory and linguistics, languages are structured following the Chomsky hierarchy, which classifies languages by the complexity of their grammatical structure \citep{Chomsky:56}. 
This hierarchy defines four types of grammars (Type-3 to Type-0) and separates formal languages by their grammatical complexity. 
A grammar in this context consists of terminal and non-terminal symbols, production rules, and a designated \emph{start symbol}, enabling the generation of valid strings within a language. 

We define a NeSy engine as a mapping $\mathcal{V}_{\mathcal{S}^*} : \Omega \times N \times \mathcal{T} \rightarrow \Sigma$, where $N \subset \Sigma$ is a set of non-terminal symbols, $\mathcal{T} \subset \Sigma$ is a set of terminal symbols and $N \cap \mathcal{T} = \emptyset$, and $\mathcal{S}^* \in \Sigma$ is a starting symbol. 
We further formalize a grammar $G = (N, \mathcal{T}, P, \mathcal{S}^*)$ with production rules defined as a $P := \mathcal{V}_{\mathcal{S}^*}(\omega, N, \mathcal{T})$.
This grammar describes the generation of symbols through expressions $\omega$.
For simplicity, we will drop the subscript of $\mathcal{V}_{\mathcal{S}^*}$ and use it as $\mathcal{V}$.
\begin{figure}[h!]
    \centering
    \includegraphics[width=1.0\linewidth]{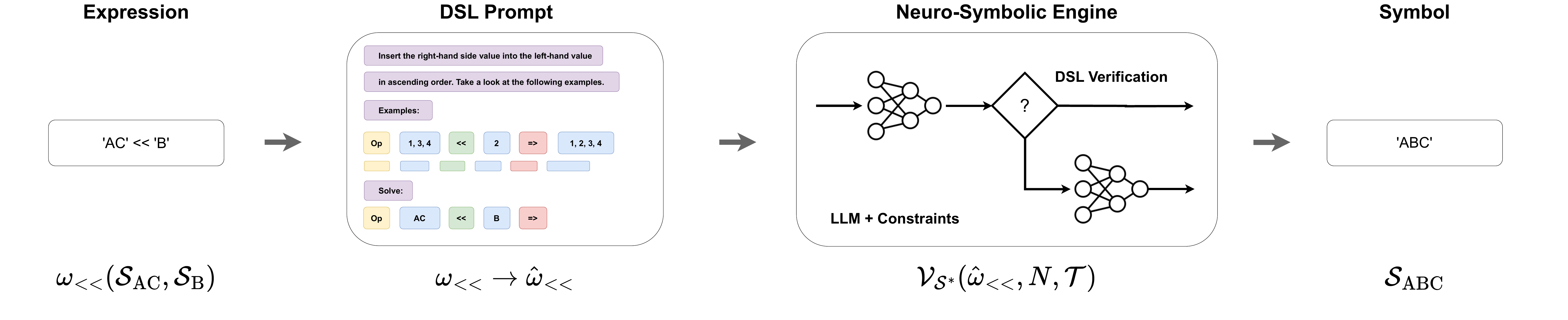}
    \caption{
    Illustration for NeSy pipeline, showcasing conceptual usage of in-context learning methodologies, domain-specific language (DSL) structures, and the expression evaluations through a NeSy engine based on an LLM and constraint verification.
    The expression showcases the sorted insert operator $\ll$ and how the information of the symbol $\text{B}$ is included in the symbol $\text{AC}$. 
    The violet placeholder in the \emph{DSL Prompt} represents an instruction, such as "\emph{Insert the right-hand side value into the left-hand value in ascending order.}" The positions below represent task-specific few-shot examples. The DSL Prompt receives the expression $\omega_{<<}$ and maps it to $\hat{\omega}_{<<}$ that can be processed by the LLM-based NeSy function $\mathcal{V}_{\mathcal{S}^*}$ and outputs a new symbol.
}
    \label{fig:prompt_engine_interaction}
\end{figure}
We identify LLMs as promising candidates for functioning as part of NeSy engines. 
In SymbolicAI, a symbol $\mathcal{S}$ is augmented with conditional instructions and types derived from DSLs, custom defined or not (e.g. HTML, SQL, etc.), tailored for directing the LLMs.
The key advantage of LLMs over previous systems lies in their ability to generalize across formal languages \citep{Wang:23} and knowledge systems.
Although there is currently no universal consensus regarding the precise classification of natural language within the Chomsky hierarchy, our approach can be understood as employing a \emph{situation-specific}, context-sensitive grammar, which enables the processing of instructions and analogies with a nuanced understanding of language.
The intersection between formal and natural languages becomes evident when considering how language patterns, through prompts like "\emph{You are a helpful assistant...}", elicit structured responses, indicating a potential underlying formal mechanism at play. 
This observation underlines the utility of such a grammar in our framework, where it serves as an explicit schema guiding the structure of examples for in-context learning.
For instance, equating "\emph{3.1415…}" with "$\pi$" or "\emph{August 4, 1961}" with "\emph{1961-08-04}" in a given context demonstrates context-dependent interpretation of symbols. 
Such a system doesn't rigidly adhere to standard grammatical rules but instead adjusts and interprets based on the  context, effectively creating a situation-specific grammar, capable of forming \emph{Domain-Invariant Associations} through in-context learning. 
We further address this in a later paragraph.

\paragraph{Function Composition}
In SymbolicAI, we use function composition to construct complex hierarchies and behaviors from fundamental elements.
Therefore, our framework enables modeling of interconnected processes, where the output of one function is used as input for another, thus creating a sequence of operations.
Through function composition, we construct computational graphs, in which intermediate symbols represent the nodes or states within these graphs.
Formally, function composition is denoted by \( \circ \), where combining functions \( f \) and \( g \) yields a new function \( h = g \circ f \), defined as \( h(x) = g(f(x)) \)
For functions \( f : X \rightarrow Y \) and \( g : Y \rightarrow Z \), their composition results in a function mapping elements from domain \( X \) to codomain \( Z \) through \( g(f(x)) \).
Although traditionally the codomain of the inner function \( f \) aligns with the domain of the outer function \( g \), SymbolicAI relaxes this constraint by allowing for any subset relationship between these domains and codomains, which is particularly beneficial for in-context learning.
When using LLMs for NeSy production rules $\mathcal{V}$, we can derive a multi-step generative process by composing a computational graph as a sequence of zero- and few-shot function compositions:
\begin{equation}
    \mathcal{V}(\omega_j, N, \mathcal{T}) = \mathcal{V}(\omega_{j-1}, \cdot) \circ \mathcal{V}(\omega_{j-2}, \cdot) \circ \dots \circ \mathcal{V}(\omega_0, \cdot),
\end{equation}
where $\omega_0$ is the initial instruction and $j$ defines the index variable for a multi-step generative process.
By leveraging functional in-context learning, where zero- and few-shot examples act as dynamic elements of the function's domain, SymbolicAI has the ability to interpret and respond to diverse input contexts.
For instance, a function can classify a user request and select an appropriate interface (e.g. WolframAlpha) to process the request. The output modality may even vary based on the respective engine.
This enables SymbolicAI to handle operations over multi-modal data that connects multi-step generative processes and establishes function composition as a central tenet in bridging multiple modalities and coordinating a variety of tasks.

\paragraph{Domain-Invariant Associations}
In-context learning enabled LLMs to become versatile task solvers by interpolating within the training distribution, to the extent that even potentially unseen tasks are addressable \citep{Brown:20}. 
We attribute this to associations formed within the input space and the capacity of Transformer architectures for defining domain-invariant feature sub-spaces.
This phenomenon has parallels with few-shot learning approaches such as SubGD \citep{Gauch:22}, a method based on identifying and utilizing a low-dimensional subspace, learned from various tasks that effectively regularize the learning process.
Since LLMs have been trained on different domains and tasks, which also include formulations of mathematical expressions, we posit that specific tokens, such as the equality sign, can be leveraged to associate meanings between different symbolic objects.
Unlike domain-invariant representations that create invariant features across different learning tasks, 
our approach leverages the in-context generalization capability of LLMs to construct invariant symbolic associations that aim to preserve, manipulate and propagate situational context.
We can use these properties to build operations that apply transformations on objects that are substitutes to the semantically aligned few-shot learning examples.

\section{SymbolicAI Framework}
\label{sup:framework}

In this section, we discuss the specifics of the proposed SymbolicAI framework. 
For more details about the framework structure, see Appendix Section~\ref{sup:framework_structure}.
For installation and usage of our framework, see Appendix Section~\ref{sec:installation}.
For more technical details and code snippets, see Appendix Section~\ref{sup:impl_details}.

\paragraph{Types and Representations}

Analogous to the type \texttt{object} in Python, the base type of SymbolicAI is a symbol represented by the base type \texttt{Symbol}. 
All other subtypes, such as \texttt{Expression}, represent their mathematical namesake and can be evaluated and simplified.
These subtypes inherit from \texttt{Symbol} the base attributes, primitive operators, and helper methods.

Although SymbolicAI uses a language-centric design, modeling and manipulating every interaction into symbolic representations is not inherently efficient.
Therefore, we establish mappings between symbolic and sub-symbolic representations for sensory inputs and non-discrete elements. 
Such mappings are typically realized through function approximation.
This allows us to map between \emph{modality}-to-language and language-to-\emph{modality} use cases.
Here, \emph{modality} serves as a placeholder for various types such as text, image, video, audio, motion, etc.
In turn, each \texttt{Symbol} object contains valued and vector-valued representations, obtained through \texttt{value} and \texttt{embedding} attributes.
The latter represents a symbol's current value, akin to embedding text and storing it as a PyTorch tensor \citep{Paszke:19} or NumPy array \citep{Harris:20}. 
While for an LLM, the numerical tensors may lack inherent meaning, vector-valued representations play an important role when 1) composite symbols are combined into more complex expressions, and 2) these embedded tensors are updated through gradient-based optimization. 

To enable the processing of symbols by LLMs, we assume that each \texttt{Symbol} object implements Python's native string functionality, where the \texttt{\_\_str\_\_} method returns an interpretable string representation.
Therefore, we can assert that any Python object is parsable by an LLM, however, the user must ensure a meaningful representation.
For more details, see Appendix Section~\ref{sup:impl_details}.

\paragraph{Polymorphic Context}

\begin{figure}[h!]
    \centering
    \includegraphics[width=1.05\linewidth]{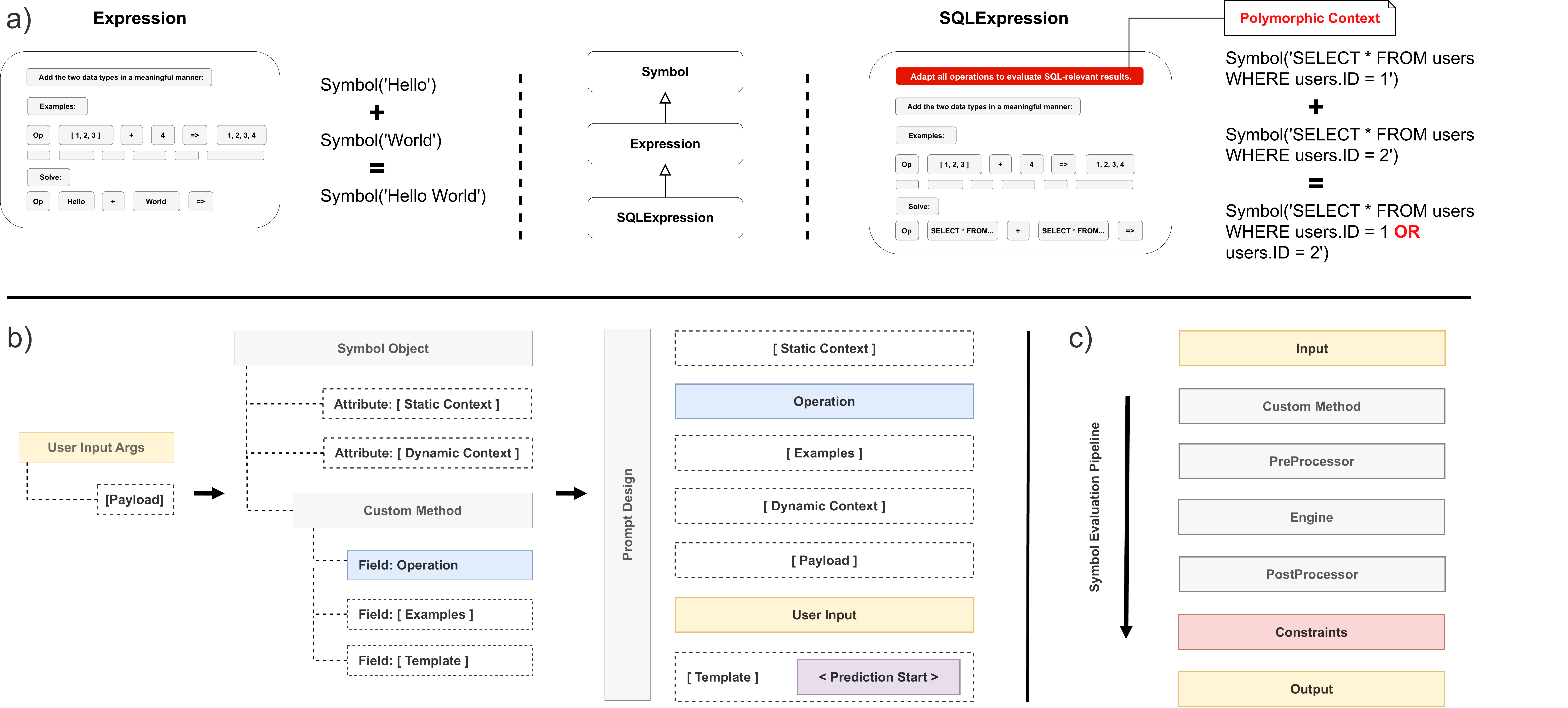}
    \caption{\textbf{a)} Illustration of polymorphic context on the example of a \texttt{SQLExpression} type for the \emph{add}-operator. Without a polymorphic context a regular \texttt{Expression} evaluation concatenates two \texttt{Symbol} objects together. The polymorphic context in \texttt{SQLExpression} overwrites the base behavior such that two added SQL-expressions get semantically combined, not concatenated. \textbf{b)} Illustration of the translation of a \texttt{Symbol} object to a prompt statement to be processed by an LLM in the NeSy engine. The \texttt{User Input Args} can be attached with a \texttt{Payload} from previous executions and gets applied to the \texttt{Custom Method}. The user input with the polymorphic context of the \texttt{Symbol Object} attributes (\texttt{Static Context} and \texttt{Dynamic Context}) are translated to a prompt statement according to the schema of the \texttt{Prompt Design}. The fields \texttt{Operation}, \texttt{Examples} and \texttt{Template} mark operation description, DSL-based prompt examples and template structures respectively. These translations are processed according to \texttt{PreProcessor} and engine-specific formatting. \textbf{c)} Illustrates the evaluation pipeline from user input to output, with multiple translation processes before and after the \texttt{Engine} invocation. The \texttt{Input} gets passed to the \texttt{Custom Method} and reformatted according to a \texttt{PreProcessor} to adhere to DSL-specific structure. The engine then takes the output of the \texttt{PreProcessor} and composes the final prompt according to the engine-specific \texttt{Prompt Design} and resolves polymorphic context and auxiliary fields. The output of the \texttt{Engine} then can be restructured by a \texttt{PostProcessor} to match DSL-requirements of the desired \texttt{Output} and gets applied \texttt{Constraints} to verify the outcome.
    }
    \label{fig:polymorphic_example_sql}
\end{figure}

Polymorphism is a central concept in programming language theory and prominently featured in SymbolicAI.
Polymorphism refers to the ability of different objects to be accessed through the same interface, or of a single identifier to represent different types based on the context of execution. 
Providing a single interface for entities of different types allows operations to be performed in ways specific to their derived types.
We therefore designed the \texttt{Symbol} object to contain a global context, which is composed of static and dynamic context parts, and enables this polymorphic behavior. 
The static context is class dependent and defined at design time.
The dynamic context is runtime adaptable and can be changed to adhere to runtime specific logic and changes.
Moreover, \texttt{Symbol} associated operations are resolved following polymorphic design before being evaluated by the NeSy engine.
SymbolicAI's engine implementation contains a \texttt{prepare} method to resolve and compile the engine specific representation by evaluating the \texttt{Symbol}-specific operations and context.
For an example on polymorphic context see part \textbf{a)} in Figure~\ref{fig:polymorphic_example_sql}.

\paragraph{Operators and Methods} 
In SymbolicAI, operators are overloaded to facilitate transformations of \texttt{Symbol} objects. 
These operator primitives employ dynamic casting to ensure type compatibility. 
Consequently, \texttt{Symbol} objects can be easily manipulated through type specific attributions or symbolically evaluated by the NeSy engine. 
For example, a central operation for boolean logic is measuring equality between symbols.
To evaluate the equality of symbols, we primarily adhere to the type specific implementation, because we prioritize strict comparisons over probabilistic evaluations. If the evaluation was unsuccessful, we then consider semantic equality through the NeSy engine.
SymbolicAI leverages decorators for composing operators and custom class methods.
For more details, see Appendix Section~\ref{sup:framework_structure}.

Upon invoking an operator or method, the respective primitive function evaluates the symbol's specific type and its respective attributes, and if necessary, resolves a nested decorated function that then uses the NeSy engine for evaluation.
Should the evaluation fail, a predefined fallback implementation executes. 
Absent a fallback, or if both evaluations fail, an error state is raised. 
The processing of an operator or custom method involves a pipeline consisting of pre- and post-processing steps, as well as constraint enforcement. 
Constraints cover aspects like return types, value ranges, and structural integrity (e.g. JSON formatting through grammar-based verification).
In Figure~\ref{fig:polymorphic_example_sql} \textbf{b)} we give an overview of the entire prompt composition based on the user input, the \texttt{Symbol} object structure, and in part \textbf{c)} the \texttt{Symbol} evaluation pipeline.

\begin{figure}[t]
    \centering
    \includegraphics[width=1.0\linewidth]{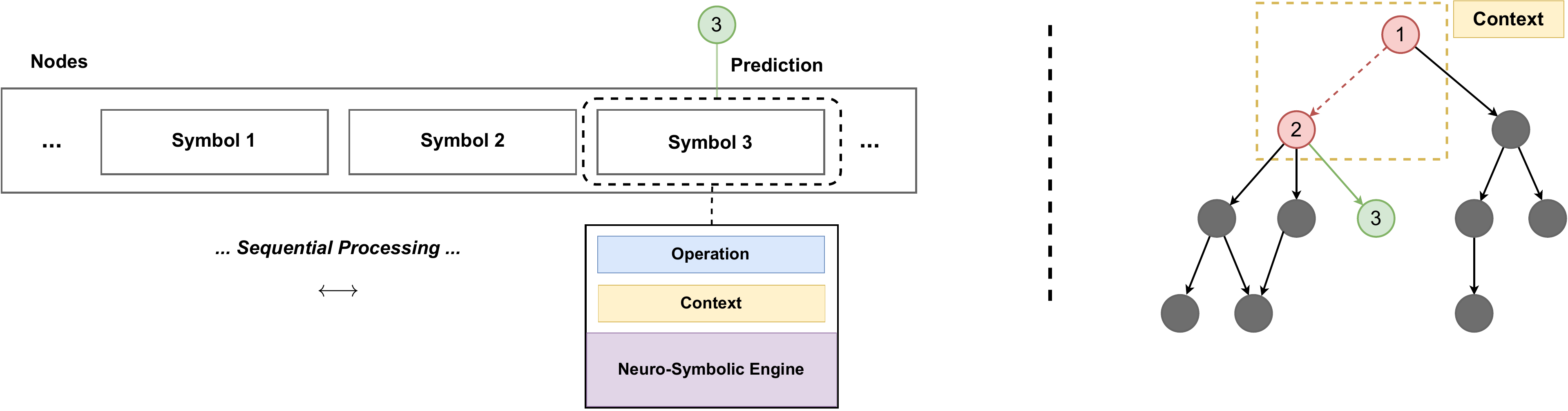}
    \caption{We showcase a multi-step hierarchical computational graph, with each node in the graph represented by a symbol. The edges are relations between symbols. The left-hand side illustrates how a new node (\texttt{Symbol 3}) is obtained by evaluating an operation with its respective context on a NeSy engine. The right-hand side illustrates the context information window (yellow rectangle) and relationship of the resulting graph with its respective nodes.}
    \label{fig:compute_graph}
\end{figure}

\paragraph{Self-Referential Structures}
SymbolicAI augments the generative process by enabling systems to introspect and modify their behavior dynamically. 
We leverage LLMs to execute tasks based on both natural and formal language instructions, adhering to the specified user objectives and with innate self-referential structures. 
We derive subtypes from \texttt{Expression} and enclose their functionalities in task-specific components, which we then expose again through templating and the model-driven design of the NeSy engine.
This design choice allows a system to create and utilize its own sub-process definitions, analogous to concepts discussed in \cite{Schmidhuber:07, Schmidhuber:09}. 
Concretely, we utilize generalization properties of LLMs to interpret and formulate a set of operations that incorporate \emph{self-instructions} \citep{Wang:22Instruct}. 
Consequently, the operations hold the flexibility to adapt to the context, and derive sub-processes that self-instruct LLMs to engage in situational modeling and context-sensitive problem-solving.
Ultimately, this enables the construction of hierarchical computational graphs for self-referential \emph{meta-reasoning} systems without the need to explicitly training a meta-learner \citep{Kirsch:22}.
In Figure~\ref{fig:compute_graph} we illustrate a step-wise evaluation of a contextual computational graph, in which the NeSy engine is processing conditioned on the current execution context and producing a next symbol prediction.

\section{Performance Measure}
\label{sec:evaluation}

One of the challenges when creating multi-step generative processes with LLMs as part of NeSy engines relies on model evaluation and handling irrelevant predictions.
The na\"ive assessment that measures only task succession would score all models to zero and render them as unusable.
Even if models follow instructions and produce parts of the expected solution, we regularly observe that they --- especially open-source models --- append a continuation of task irrelevant predictions. 
Such predictions result in failure modes when applying conditions and validations, and halt any multi-step procedure.
Our solution is an evaluation protocol that refines the performance measurement, allowing for more nuanced diagnostics and the possibility of continuing the evaluation despite intermediate failures.
To derive our quality measure, we borrow ideas from the utilization of the Fréchet distance for generative processes \citep{Heusel:17}.

We generate trajectories through a NeSy sequential process that creates a trajectory of distributions $\mathbb{P}$ over multiple iterations of generative nodes.
Each node in the process can be aligned to a reference distribution, which marks the desired behavior. 
To quantify the validity of the generated trajectories, we measure the total distance between the generated and reference data distribution along the path trajectory. 
We therefore adopt a cumulative measure capable of taking into account the entire generative trajectory.
In theory, this process would entail calculating the path integral over the latent space representations for models, cumulating the Fréchet distances \citep{Dowson:82} traversed along these trajectories:
\begin{equation}
\mathcal{D}(\mathbb{P}_{\text{gen}}, \mathbb{P}_{\text{ref}}) = \int_{t_0}^{t_f} d(\mathcal{N}(m_t, C_t), \mathcal{N}(m_{w,t}, C_{w,t})) \, dt
\end{equation}
where $\mathcal{D}(\mathbb{P}_{\text{gen}}, \mathbb{P}_{\text{ref}})$ denotes the integral of the Fréchet distances between two data distributions along the generative path trajectory from an initial time $t_0$ to a final time $t_f$, $d(\mathcal{N}(m_t, C_t), \mathcal{N}(m_{w,t}, C_{w,t}))$ is the Fréchet distance calculated at each time $t$ between the generated multivariate normal data distribution with mean $m_t$ and covariance $C_t$, and the reference multivariate normal data distribution with mean $m_{w,t}$ and covariance $C_{w,t}$.
The resulting measure follows properties of normal distributions and is consistent with increasing disturbances.

However, this approach is computationally intractable for large-scale problems, and requires access to latent representations, which --- especially in the context of LLMs --- is not always given.
For computational feasibility, we introduce an approximation that measures the embedding distances over the path trajectories through an auxiliary embedding model, based on prior work on distribution regression \citep{Szabo:16}.
The embedding model maps the symbolic representations into a RKHS, such that we can apply a kernel mean embedding function to measure their respective distances \citep{You:19, Dinu:23}.
We assess the distance through the mean embeddings w.r.t. to a kernel function $K(\cdot, \cdot)$ of the
samples $\mathbf{e}^t_{x} \sim \nu^t_{\text{gen}} \in \mathbb{P}_{\text{gen}}$ and $\mathbf{e}^t_{y} \sim \nu^t_{\text{ref}} \in \mathbb{P}_{\text{ref}}$ 
produced by the generated data distribution and a reference data distribution respectively.
We denote by $\mu_{\mathbf{e}^t_{x}}$, $\mu_{\mathbf{e}^t_{y}}$ the mean embeddings associated to the respective samples, i.e.  $\mu_{\mathbf{e}^t_{x}}(z)=\frac1n \sum_{i=1}^n K(x_i^t,z)$ in case $\mathbf{e}^t_{x}=(x_i^t)_{i=1}^n$ is a sample of size $n$ of the respective mean embeddings.
To compute the similarity between the embeddings of the generated and reference distributions, we evaluate the associated maximum mean discrepancy $\text{MMD}^2(\mu_{\mathbf{e}^t_{x}},\mu_{\mathbf{e}^t_{y}})$ \citep{Gretton:12} and then, as before for the Fréchet distances, we integrate over $t$:
\begin{equation}
\mathcal{\tilde{D}}(\mathbb{P}_{\text{gen}}, \mathbb{P}_{\text{ref}})=\int_{t_0}^{t_f}\text{MMD}^2(\mu_{\mathbf{e}^t_{x}},\mu_{\mathbf{e}^t_{y}})dt.
\end{equation} 
In empirical evaluations, however, we care about normalized values for ease of interpretation.
We therefore analyze the properties of the MMD and derive a similarity score, which follows the same statistical principles as the MMD, and is bound between $[0, 1]$. We concluded that we can utilize only the MMD cross terms to evaluate the similarities. See Appendix Section~\ref{sup:mmd_fd} for more details.
For our comparisons as referenced in Figure~\ref{fig:spider_plot} we therefore denote the similarities rather than distances.
We then come to the following formulation and refer to our empirical measure as the "Vector Embedding for Relational Trajectory Evaluation through Cross-similarity", or \emph{VERTEX}~score for short: 
\begin{equation}
s(\mathbb{P}_{\text{gen}}, \mathbb{P}_{\text{ref}}) := \int_{t_0}^{t_f} \big [\min(\max(0,\frac1z \widetilde{\text{MMD}^2}(\mu_{\mathbf{e}^t_{x}},\mu_{\mathbf{e}^t_{y}})-z_{\mathrm{rand}}),1 ) \big ] dt.
\end{equation} 
We approximate the integral across time steps through Monte Carlo approximation.
The introduced normalization constants denote the similarities to a random~sequence~$z_{\text{rand}}$, which functions as a baseline subtraction to recenter our results, and a given reference score to rescale w.r.t. to scores obtained from comparing related solutions~$z$.
Min-max scaling ensures the final measure is bounded between $[0, 1]$. This process reflects properties such as H\"older continuity that bounds the kernel function within certain limits. 
To compute the embeddings, we utilize the embedding model \texttt{all-mpnet-base-v2} \citep{Song:20}, due to its widespread availability, and its balance between speed and quality.
As a similarity measure, we select a Gaussian kernel following our derivation from the Appendix Section~\ref{sup:mmd_fd}.
In our implementations, we also explore other kernels, including preliminary experiments with cosine similarity. 
We also note that one can integrate Bernoulli distributed trials into our score, with $0$ values representing failure modes and values of $1$ being successes. 
Furthermore, if we relax our definition, we can integrate other similarity measures which are bound between $[0, 1]$, which then reflect on domain-specific attributions, i.e. including a similarity measure tailored towards capturing the nuances between two sub-structures of abstract syntax tree.

\section{Evaluation}
\label{sec:experimental_design}

We introduce a benchmark that evaluates multi-step generative processes as complex workflows. 
Our benchmark consists of five different evaluation categories, and uses the VERTEX score to measure the capabilities of an LLM to solve tasks from each category.
The five categories of our benchmark are divided into three basic evaluations and two advanced categories that combine different basic capabilities.
The three basic categories are (i) \textbf{associative~prediction} which measures a models proficiency in understanding associations between symbols, (ii) \textbf{multi-modal~binding} where we evaluate the capability to employ tools and operate on different modalities, and (iii) \textbf{program~synthesis} for measuring a models proficiency in generating consistent code and executing subroutines.
The two advanced benchmark categories are (iv) \textbf{logic}, for evaluating logic-based components and (v) \textbf{computational~graphs} where complex workflows need to be processed, evaluating all aforementioned capabilities.
For our evaluation we focus on the GPT family \citep{Brown:20} of models, specifically GPT-3.5~Turbo (revision 1106) and GPT-4~Turbo (revision 1106) as they are the most proficient models to date; Gemini-Pro \citep{Google:23} as the best performing model available through API from Google; LLaMA2-Chat~13B \citep{Touvron:23}, LLaMA3-Chat 8B and LLaMA3-Chat 70B from Meta represent open-source LLMs. Finally, Mistral 7B \citep{Jiang:23} and Zephyr 7B \citep{Tunstall:23} serve as baselines for revised and fine-tuned open-source models. 
The open-source models Mistral, Zephyr, and smaller LLaMA variants are estimated to have roughly equivalent parameter counts compared to GPT-3.5~Turbo and Gemini-Pro.
All our experiments require a context size smaller or equal to $4096$ to enable the comparisons among the in-context capabilities across model architectures.
For the LLaMA models, we utilize the \emph{chat} versions since they are specifically tuned to follow instructions.

\paragraph{Associative Prediction}
We evaluate a model's proficiency to follow simple and complex instructions and associations with zero- and few-shot examples. Therefore, we evaluate the proficiency in applying our operators between \texttt{Symbol} types. We defined a total of 15 tasks involving in-context associations between two \texttt{Symbol} instances. SymbolicAI's overloaded operators rely on predefined pseudo-grammars, as described in Section \ref{formal_languages}, that augment the operators with few-shot examples. For instance, the overloaded operator \texttt{+} utilized between two \texttt{Symbol} instances provides few-shot examples how to resolve additions with various data types.
Consequently, we can now test if the models can solve the addition between \texttt{Symbol("two hundred and thirty four")} and \texttt{Symbol(7000)}. See Appendix Section~\ref{sup:associative_pred} for more details.

\paragraph{Multi-modal Binding}
We perform transformations between multiple modalities through language-based representations.
Therefore, we need to evaluate the model's proficiency in tool utilization, classification and routing of requests to relevant modules. 
We define a multi-modal \ttt{Expression} to detect the category of a task based on its content and to forward the task to the appropriate tool. The expression creates interfaces to tools like WolframAlpha for mathematical expressions, Selenium for website content scraping, SerpApi for search queries, and APILayer for optical character recognition.
Each of the five tests aims to evaluate the appropriate handling of a specific type of input by the multi-modal \ttt{Expression} type, such as processing a website URL for scraping, interpreting a search engine query, testing if two vectors are linearly independent, comparing large numbers, and extracting text from an image.
See Appendix Section~\ref{sup:multimodal_binding} for more details.

\paragraph{Program Synthesis}
We evaluate executable code with and without concepts from retrieval augmented generation, model-driven development, and experiment with self-generated instructions by creating self-referential expressions. 
We designed three separate tests related to program synthesis, where each task assesses the ability of the models to generate and execute code based on natural language instructions or provided templates:

1) The first task involves reading a LaTeX table template and data, then generating a function to populate the table with the given data. 

2) The second task tests the automatic code generation for API calls by fetching data from a specified URL and extracting specific information from the retrieved content. 

3) The third task evaluates the ability to construct a custom \ttt{Expression} that processes a \ttt{Symbol} through a specific \ttt{Function} component from the SymbolicAI package. 

Each of the three tests follows a similar pattern, where the generated code is scored based on its similarity to valid references and normalized with random samples.
See Appendix Section~\ref{sup:program_synthesis} for more details.

\paragraph{Logical Components}
To evaluation the capabilities for logical reasoning of models, we condition them to create a sequence of expressions as self-contained components, and refer to higher-order logic for their assessment. 
Based on the underlying \emph{type theory} originating from \cite{Whitehead:1925}, we evaluate a models' capability to resolve statements in the form of \emph{there exists x such that x fulfills y}.
Such quantifiers define the standard semantics of expressions, where their meaning is given by a semantic function.
A semantic function maps a term from an abstract definition to a point in a domain, which is an interpretation of the term's type and value.
Therefore, these functions operate on types and values of expressions, and relations thereof.
Subsequently, NeSy engines can formulate and evaluate at inference time logic-based instructions through Lisp, Prolog, or Mathematica \citep{McCarthy:59, Colmerauer:93, Chen:93, Mathematica:22}, or leverage solvers such as Z3 \citep{Moura:08}. 
Therefore, the result of a natural language statement when evaluated by a NeSy engine can be interpreted by any expert system which defines the corresponding semantic functions and process them either in a symbolic \citep{Feigenbaum:65, Rose:94}, differentiable \citep{Velickovic:21, Ibarz:22}, or hybrid manner \citep{Kuncicky:91}.

We evaluate how proficient models are at interpreting custom DSLs and define expression statements. 
DSLs are designed to express logical relations and operations in a structured format, and supports human-readable and machine-interpretable formulations. 
The following example illustrates such relationships by translating a natural language statement into an expression statement, as follows: 
\begin{quote}
Marvins has four paws and likes to meow when I pet its fur. Is Marvins a cat?
\end{quote}

A DSL may enforce the usage of $\text{HAS}(\cdot)$, $\text{IS}(\cdot)$, etc. and may condition an LLM to produce the following expressions:
\begin{itemize}
    \item $\text{HasFourPaws}(x)$: $x$ has four paws.
    \item $\text{LikesToMeowWhenPetted}(x)$: $x$ likes to meow when it is petted.
    \item $\text{IsCat}(x)$: $x$ is a cat.
\end{itemize}

These are then utilized to define the following logical expression:
\begin{equation*}
    \forall x \big (\text{HasFourPaws}(x) \land \text{LikesToMeowWhenPetted}(x) \Rightarrow \text{IsCat}(x) \big).
\end{equation*}

An automated theorem prover can now evaluate this statement for all values of $x$ and assess the validity of the original query.  
Lastly, our evaluation uses symbolic mathematics to manipulate algebraic expressions. 
This involves defining symbols and performing operations like factorization, simplification, and algebraic manipulation. 
The symbols are placeholders for any value, enabling the definition of general expressions without specifying their values upfront. 

We designed six tests to assess the logical capabilities of the candidate models and group them as follows. See Appendix Section~\ref{sup:logical_comps} for more details.

1) We utilize the Python library SymPy for symbolic mathematics to create the mathematical expression $ax + bx - cx - ay - by + cy + d$. The task for the model is then to factorize the expression and extract all unique symbols as a list.

2) Three tasks evaluate a models' capability to resolve the logical operations AND, OR, and XOR. 
For instance, the test for logical AND combines the symbols \ttt{Symbol("The horn only sounds on Sundays")} and \ttt{Symbol("I hear the horn")} and compares the answer against the human-generated references "\textit{The horn only sounds on Sundays and I hear the horn.}" and "\textit{Since I hear the horn it is Sunday.}" 
Since there is a large number of possible solutions, there is high variability in the solution space. Each model might prefer a different solution. 

3) For another task we use a custom \ttt{Expression} that defines a DSL syntax and semantic structure.
We use this Expression to extract higher-order logic expressions from a natural language statement, namely the puzzle 'Who is Jay's brother?'\footnote{\ Bob has two sons, John and Jay. Jay has one brother and father. The father has two sons. Jay's brother has a brother and a father. Who is Jay's brother?}, that preserves the original relationships. 

4) For the final task, we again use the puzzle 'Who is Jay's brother?' to evaluate a models' capability for complex conversions. 
We use the Z3 theorem prover \citep{Moura:08} to solve the 'Who is Jay's brother' puzzle conditioned on the Z3 solvers' solution to Einsteins' famous puzzle 'Who owns the fish?'.
The task involves an indirect translation from natural language to executable code by the Z3 solver; the solution to Einstein's puzzle acts as a form of self-contained "documentation" for how the Z3 solver should be utilized.
The test constructs a template, which includes the task instructions, puzzle statement, and reference to the Einstein's puzzle solution. 
The models are then asked to analyze the given problem and solution format and create a Python function with Z3 syntax that can solve the 'Who is Jay's brother?' puzzle.
The dynamically generated code is executed within the test environment utilizing Python's \ttt{exec} function. 
We check the access to the Z3 solver and run the generated \ttt{solve\_puzzle} function supposed to contain the logic to solve the puzzle.
Once executed, the assembled Z3 logical clauses are processed by the solver, which verifies that the set of constraints is satisfiable. 
If so, the model generated by the solver is queried for the puzzle's solution and scored using our VERTEX score.

\paragraph{Hierarchical Computational Graphs} \label{hierarchical_computation_graphs}

\begin{figure}[h!]
    \centering
    \centering
  \begin{minipage}[c]{0.55\textwidth}
    \begin{adjustbox}{max width=\linewidth}
    \includegraphics[width=0.92\linewidth]{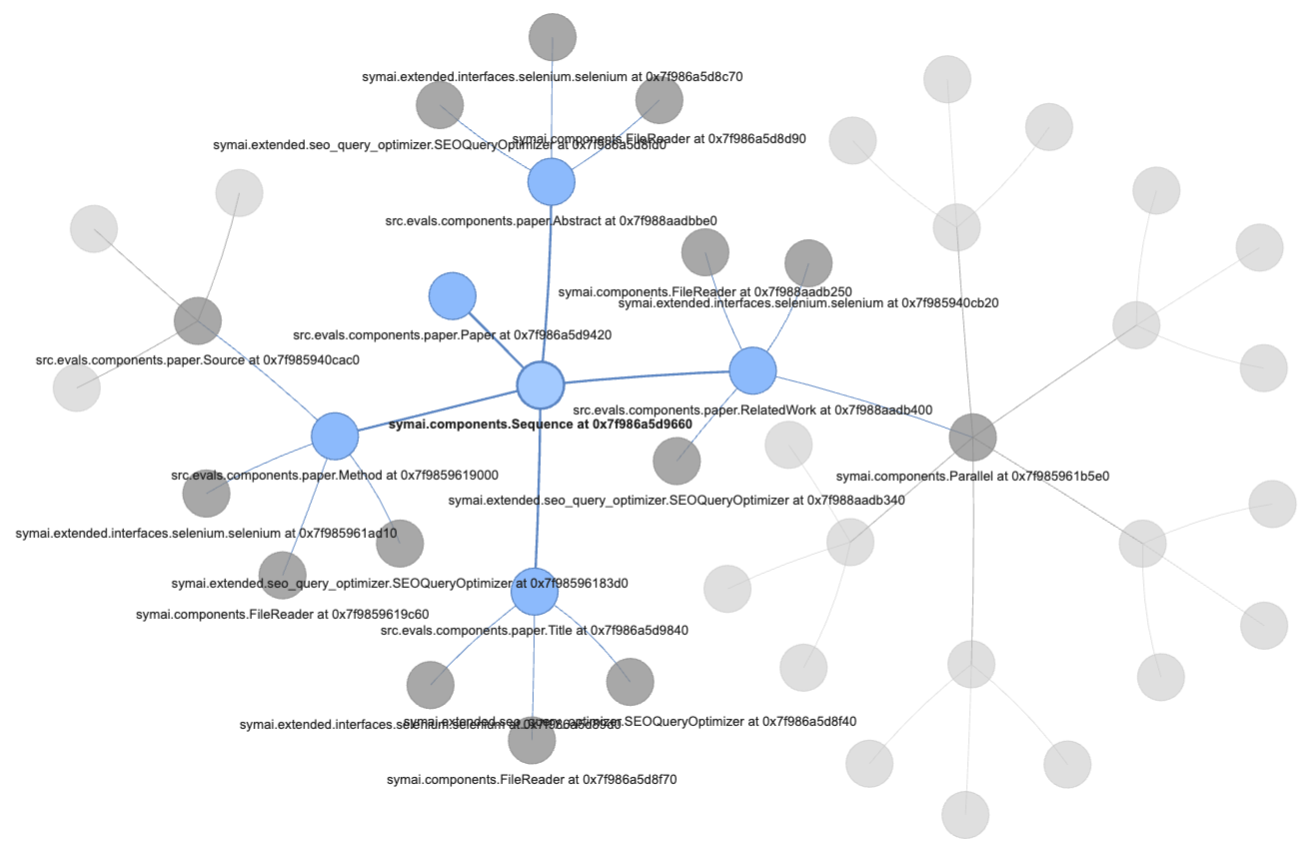}
    \end{adjustbox}
  \end{minipage}\hfill
  \begin{minipage}[c]{0.45\textwidth}
    \caption{We illustrate the hierarchical computational graph for the \texttt{Paper} expression. Each node represents an instance of an expression with distinct properties and behaviors, such as file sourcing, generative process, tool utilization, or transformation operation. The edges denote the reference relationships between expressions and indicate the flow of information. The blue highlighted nodes mark the main sequence nodes of expressions utilized to create parts of the paper draft, such as \texttt{Method} section, \texttt{RelatedWork} section, \texttt{Abstract} section, and so on. Each generative node is used for evaluating the VERTEX score. None-generative nodes such as search engine results are not evaluated, and we assume to obtain ground-truth values.}
    \label{fig:graph_result}
  \end{minipage}
\end{figure}

We evaluate the capabilities of models to orchestrate a multi-step generative process and evaluate a set of tasks. 
Models need to direct sub-processes and associate computational results from and to \texttt{Symbol} nodes, and maintain relationships between these nodes, which we refer to as a computational graph as shown in Figure~\ref{fig:graph_result}. 
In a computational graph, the VERTEX score compares the results produced by a generative model at each node against samples obtained from a reference distribution, usually modeled by sampling from multiple valid references.
We also account for randomness through predefined random samples for normalizing the result.
Our reference to \emph{hierarchical} computational graphs stems from the fact that we operate on multiple levels.
On a higher level of abstraction we are able to perform planning, sub-task scheduling, and define operational instructions. 
On a lower level of abstraction, we execute these plans based on the defined instructions and data, which can also span generative processes that produce new information.

Given that the field is currently at an early stage in developing even sequential schedulers for LLM-based planning systems, our evaluations will be confined to sequential execution only.
We introduce two tests designed to evaluate multi-step generative processes:

1) We simulate and evaluate the process of writing a research paper draft based on a predefined hierarchical computational graph that focuses on the content output of the computational graph rather than planning and scheduling functionality. See Appendix Section~\ref{sup:hcg} for more details.

2) We test the VERTEX Protocol as defined in Algorithm~\ref{algo:evaluation_protocol}, which represents our general method for evaluating multi-step generative processes. 
We create a self-contained test scenario to illustrate an end-to-end evaluation and as a go-to reference for how our protocol can be deployed in a realistic environment. Our evaluation protocol is not only designed to analyze and score a series of instructions, but also to provide a structured basis for recording these processes. 
Furthermore, we note that our evaluation protocol is generally formulated, which allows the application of non-sequential planning and scheduling.

\algnewcommand{\algorithmicgoto}{\textbf{go to:}}%
\algnewcommand{\Goto}[1]{\algorithmicgoto~\ref{#1}}%
\algnewcommand\algorithmicinput{\textbf{Method:}}
\algnewcommand\Method{\item[\algorithmicinput]}

\begin{figure}[ht]
\begin{algorithm}[H]
\caption{VERTEX Protocol}
\begin{algorithmic}[1]
    \Require NeSy engine $\mathcal{V}$ as an LLM, embedding engine $\mathcal{E} : \Sigma \rightarrow \mathcal{H} \subset \mathbb{R}^d$, symbols $\{x_0, x^{*}, y^{*}\} \subset \Sigma$, with $x_{0}$ as the initial instruction, $x^{*}$ as the payload resulted from executing $\mathcal{V}$, $y^{*}$ as the reference, and $*$ acting as a placeholder for $\mathcal{P}, \mathcal{T}, \mathcal{C}$, capabilities $\mathcal{C} = \{\mathcal{F}_1, \mathcal{F}_2, \mathcal{F}_3, \ldots\}$, where each $\mathcal{F}_i$ represents a specific functional role within the system, plan $\mathcal{P} \subset \Sigma$, task $\mathcal{T} \in \mathcal{P}$, memory buffer $\mathcal{M} \subset \Sigma$, a scoring function $\tilde{s}:\mathcal{H} \times \mathcal{H} \rightarrow[0,1]$, a scheduler $\mathcal{Q}$, an aggregator $\mathcal{A}$, and score variables $\{s \} \in [0,1]$.

    \Method 
    \State $\mathcal{V}, \mathcal{E}, \mathcal{Q}, \mathcal{C}, y^{\mathcal{P}} \leftarrow$ \Call{Init}{$\cdot$} \Comment{Initialize engines, scheduler, capabilities, expected plan.}
    \State $\mathcal{M} \leftarrow \emptyset, \mathcal{A} \leftarrow \emptyset$ \Comment{Initialize memory buffer and aggregator.}
    \State $x^{\mathcal{P}} \leftarrow$ \Call{GeneratePlan}{$x_0, \mathcal{V}$} \Comment{$\mathcal{V}$ generates plan based on initial instruction.}
    \State \Call{Evaluate}{$x^{\mathcal{P}}, y^{\mathcal{P}}, \mathcal{E}, \mathcal{A}, \tilde{s}$} \Comment{Embed, score, and aggregate plan similarity.}
    \State $\mathcal{P}, \mathcal{M} \leftarrow$ \Call{UnfoldPlan}{$y^{\mathcal{P}}, \mathcal{M}, \mathcal{Q}$} \Comment{$\mathcal{Q}$ unfolds plan into actionable tasks and updates progression.}
    
    \While{$\mathcal{P} \neq \emptyset$} \Comment{Run until list of tasks is exhausted.}
        \State $\mathcal{T}, y^{\mathcal{C}}, y^{\mathcal{T}} \leftarrow$ \Call{Select}{$\mathcal{M}, \mathcal{V}$} \Comment{$\mathcal{V}$ selects next task based on task progression.}
        \State $\mathcal{F}_i \leftarrow$ \Call{Identify}{$\mathcal{T}, \mathcal{C}, \mathcal{V}$} \Comment{$\mathcal{V}$ identifies task-related capability $\mathcal{F}_i$.}
        \State $x^{\mathcal{C}}, x^{\mathcal{T}} \leftarrow$ \Call{Execute}{$\mathcal{T}, \mathcal{F}_i, \mathcal{Q}$} \Comment{$\mathcal{Q}$ executes $\mathcal{T}$ with capability $\mathcal{F}_i$ and assign results $x^{\mathcal{C}}, x^{\mathcal{T}}$.}
        \State \Call{Evaluate}{$x^{\mathcal{C}}, y^{\mathcal{C}}, x^{\mathcal{T}}, y^{\mathcal{T}}, \mathcal{E}, \mathcal{A}, \tilde{s}$} \Comment{Embed, score, and aggregate capability similarity.}
        \State $\mathcal{P}, \mathcal{M} \leftarrow$ \Call{Update}{$\mathcal{T}, \mathcal{P}, \mathcal{M}, \mathcal{Q}$} \Comment{$\mathcal{Q}$ updates plan and progression.}
    \EndWhile
    
    \State $s$ $\leftarrow$ \Call{Finalize}{$\mathcal{A}$} \Comment{Finalize aggregation of scores.}
    \State \Return $s$ \Comment{Return aggregated score of plan execution.}

\end{algorithmic}
\end{algorithm}
\setcounter{algorithm}{0}
\vspace{-1.5em}
\captionof{algorithm}{This algorithm defines the pseudocode of our VERTEX protocol with our respective VERTEX score as a scoring criteria. We start by initializing the NeSy engine $\mathcal{V}$, the embedding engine $\mathcal{E}$, the scheduler $\mathcal{Q}$, and a set of capabilities $\mathcal{C}$. The initial instruction $x_0$ is utilized to generate a plan $x^{\mathcal{P}}$ through $\mathcal{V}$. The plan and its expected outcome $y^{\mathcal{P}}$ are embedded, and their similarity is scored according to our VERTEX score and aggregated. The plan is then unfolded into actionable tasks. Each task $\mathcal{T}$ is selected and executed with the appropriate capability $\mathcal{C}$, resulting in the capability and task results $x^{\mathcal{C}}, x^{\mathcal{T}}$, and expected outcomes $y^{\mathcal{C}}, y^{\mathcal{T}}$ updated in the memory buffer $\mathcal{M}$. The process continues, with each task's result being embedded, scored, and aggregated until the plan is complete. The final aggregated score $s$ is returned, reflecting the overall effectiveness of the plan execution.}
\label{algo:evaluation_protocol}
\end{figure}

We start with a high-level workflow description which consists of a list of tasks and optionally their respective sub-tasks; we refer to this as the plan $\mathcal{P}$.
To perform the experiment, we utilize an expected plan $y^{\mathcal{P}}$ which was handcrafted for this evaluation. 
The expected plan is a queue of predefined tasks (in a particular order) that the system should follow to achieve the goal.
The goal statement defines the end objective that the workflow aims to accomplish.
We also have a set of plans similar to the expected plan, which are trajectories in the solution space, as well as the plan $x^{\mathcal{P}}$  that the LLM generates utilizing the \Call{GeneratePlan}{} call for a specific seed.
We score the predicted plan against the expected plan and the trajectories, then we continue to the next phase in which we utilize the expected plan to execute the tasks.
At each step, the LLM will receive in its context the goal, the tasks, the current progress, and a query asking for the next task to execute; we refer to this as the memory buffer $\mathcal{M}$.
If the LLM is not able to predict the next task, it will return a failure, and the expected plan will be utilized to execute the next task.
The LLM has access to a predefined set of capabilities $\mathcal{C}$, specifically WolframAlpha, SerpApi, Selenium, and the LLM itself, which also represents our self-referential structure. 
We keep executing tasks until the queue is exhausted, and at each step, we utilize the \Call{Evaluate}{} call to measure the performance of the LLM with our VERTEX score.
The scheduler class $\mathcal{Q}$ oversees the execution of the test workflow.
It takes the setup configuration and orchestrates the linear execution of tasks, utilizing the expected plan as a reference.
It maintains a pool of tasks to be executed and updates progress as tasks are completed.
The \Call{UnfoldPlan}{} call is a method of the scheduler class $\mathcal{Q}$.
The method calls itself recursively until there are no tasks left.
The \Call{Select}{} call is responsible for determining which task to execute next from a pool of remaining tasks.
It utilizes the LLM through self-reflection \citep{shinn2023reflexion} to choose the most suitable next task based on a template that gets progressively updated in the memory buffer $\mathcal{M}$ by the \Call{Update}{} call. 
The \Call{Identify}{} call uses self-reflection and similarity scoring to determine the best interface based on the task at hand, then passes the interface to the \Call{Execute}{} call to execute the task.
Lastly, the test ends with the \Call{Finalize}{} call, which provides an aggregated assessment of the model's ability to manage and execute the workflow.

\begin{figure*}[ht]
  \centering
  \begin{minipage}[c]{0.45\textwidth}
    \begin{adjustbox}{max width=\linewidth}
    \includegraphics[width=\linewidth]{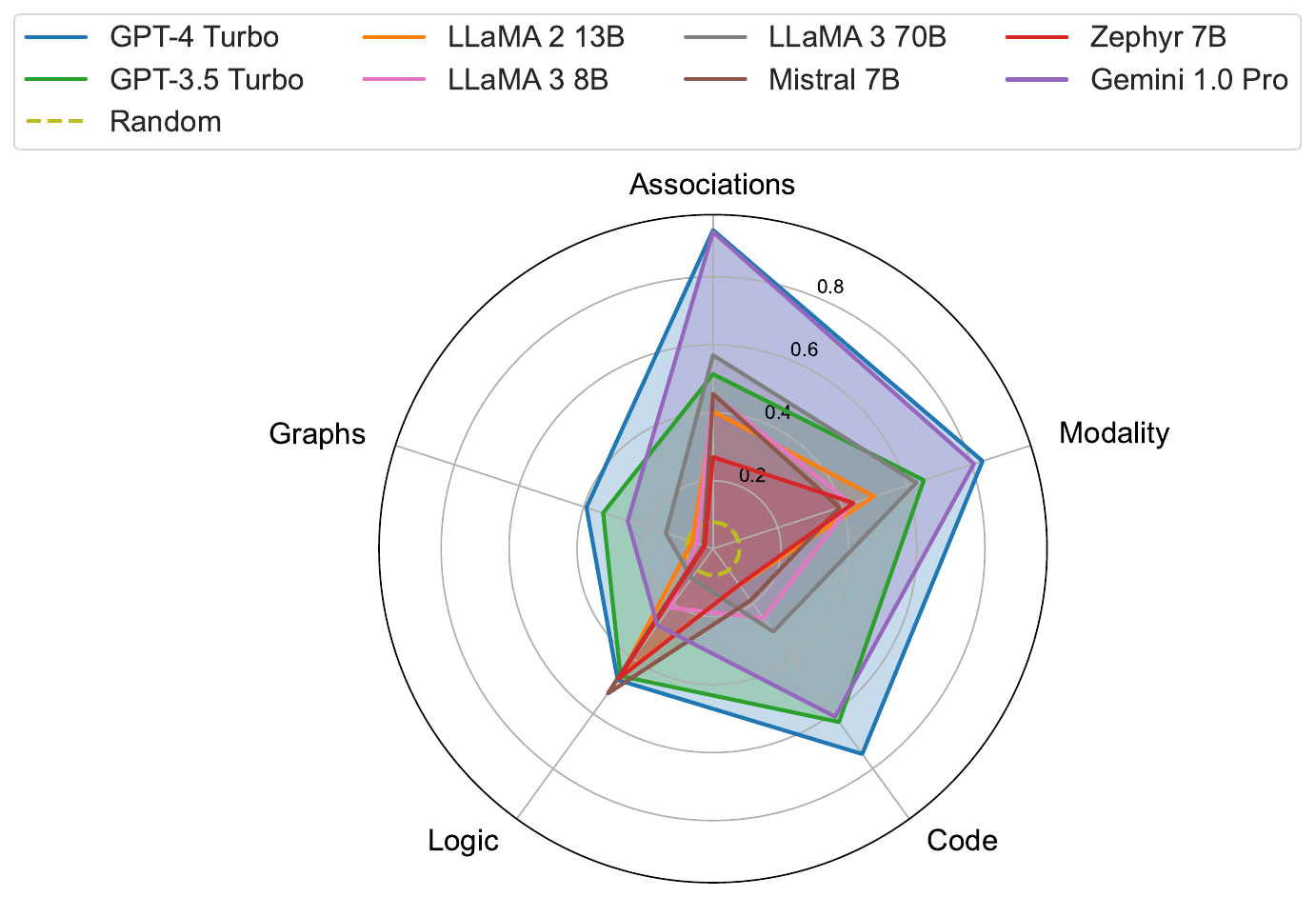}
    \end{adjustbox}
  \end{minipage}\hfill
  \begin{minipage}[c]{0.5\textwidth}
    \caption{We evaluate GPT-4 Turbo, GPT-3.5 Turbo, Gemini-1.0 Pro, LLaMA2-Chat 13B, LLaMA3-Chat 8B, LLaMA3-Chat 70B, Mistral 7B and Zephyr 7B on five benchmark categories: 1) Associative Prediction (Association) 2) Multi-modal Binding (Modality) 3) Program Synthesis (Code) 4) Functional Logic Components (Logic) and 5) Hierarchical Computational Graphs (Graphs). We denote the VERTEX scores for each category as a normalized value between $0$ and $1$, where higher values are better. The VERTEX score is measured according to a reference baseline and normalized by random sequences to  exclude noise and similarities among references distributions to rescale solutions.
    The shown scores are an average over all tests per category and across 8 different seeds per test.}
    \label{fig:spider_plot}
  \end{minipage}
  ~
  \begin{minipage}{\textwidth}
    \centering
    \begin{adjustbox}{max width=\linewidth}
    \begin{tabular}{lccccccccc}
    \toprule
      \textbf{Benchmarks} &  GPT-4 Turbo & GPT-3.5 Turbo & Gemini 1.0 Pro & LLaMA 2 13B & LLaMA 3 8B & LLaMA 3 70B & Mistral 7B & Zephyr 7B & Random \\
      \midrule
      Associations & \textbf{0.94} & 0.51 & 0.93 & 0.40 & 0.46 & 0.57 & 0.46 & 0.27 & 0.08 \\
      Modality & \textbf{0.83} & 0.65 & 0.81 & 0.50 & 0.43 & 0.63 & 0.39 & 0.43 & 0.07 \\
      Code & \textbf{0.75} & 0.63 & 0.61 & 0.13 & 0.25 & 0.30 & 0.19 & 0.13 & 0.00 \\
      Logic & 0.48 & 0.46 & 0.28 & 0.46 & 0.21 & 0.11 & \textbf{0.53} & 0.47 & 0.00 \\
      Graphs & \textbf{0.39} & 0.34 & 0.26 & 0.06 & 0.05 & 0.15 & 0.03 & 0.03 & 0.00 \\
      \midrule
      \textbf{Total} & \textbf{0.68} & 0.52 & 0.58 & 0.31 & 0.28 & 0.35 & 0.32 & 0.27 & 0.03 \\
      \bottomrule
    \end{tabular}
    \end{adjustbox}
    \label{tab:benchmark_results}
  \end{minipage}
\end{figure*}

In Figure~\ref{fig:spider_plot} we conclude with our evaluation and compute the cumulative score for all described evaluation categories and in the next section we discuss how to interpret the results of our framework.

\section{Discussion}
\label{sec:discussion}
In this section, we address the evaluation results, auxiliary findings and limitations of SymbolicAI and the future directions we are focusing on.
Some of the limitations stem from the inherent constraints of current technologies and dependencies on third-party systems. 
Additionally, the nuanced complexities of working with generative models presents further challenges.

\subsection{Results}
In Figure~\ref{fig:spider_plot} we show the VERTEX score for all five evaluation categories on 8 different state-of-the-art models.
We show the aggregated results per category, meaning the average score among all tests averaged per category and average across 8 different seeds per test.
The VERTEX score is normalized between $0$ and $1$, where higher values are better.
Our score is non-linear due to its nature of using non-linear kernels, and captures semantic, ordinal and relative structures among the data samples.
However, since our score is highly dependent on the quality of the underlying embedding model, it may omit to capture fine-grained syntactic differences such as `Hello` vs `hello`. 

In our experiments, we have noticed that for associative predictions and multi-modal bindings, GPT-4 Turbo is on par with Gemini-1.0 Pro. 
Furthermore, there is still a large gap between open-source contestants such as LLaMA 3 even with 70B parameters compared to the closed-source alternatives from OpenAI and Google.
For the rest of the experiments, we see that GPT-4 almost always outperforms all other models, except for the functional logic components category. 
Here, we analyzed results and found that the larger models sometimes take shortcuts by automatically returning the solution and answering that the task instructions are too complex for such a straight-forward puzzle query.
However, we would rather state in general that for logic-based, planning and scheduling tasks all models act unreliably, even if slight performance differences between the models are seen in the plot.
We believe this is in part due to lack of training data  specifically for workflows, planning and scheduling tasks, and to imprecision in generating reliably structured output formats, such as custom DSLs or other custom in-context instructed formats.
This also stems from their instruction fine-tuning, since most models are chat-based models and offer verbose responses which need to be  suppressed or post-processed.

We see similar performance between GPT-3.5 Turbo and LLaMA 3 70B except for the logical and graphs evaluations.
We found that LLaMA 3 70B has a tendency to ask questions back if it does not understand the request instead of following the specified instructions provided.
We assume this also stems from the chat-based instruction fine-tuning.
Zephyr 7B and Mistral 7B have shown on par capabilities in functional logic components with larger models, however fail in program synthesis and hierarchical computational graphs experiments.
We observe that they perform well when resolving the overloaded logic operators such as OR, AND and XOR, and show decent performance for text generation, but fail to resolve more complex instructions.

\subsection{Limitations}
\label{sec:limitaiton}

\paragraph{Framework}
Since the framework interfaces with many tools and API services, it requires a substantial engineering feat to integrate all available functionalities and keep the API-based services up-to-date.
For us this means, that although we support a variety of tools and frameworks like Selenium, WolframAlpha, or Z3, we only scratch the surface of these tools.
Moreover, the utilization of grammar-based constraints validations is still experimental and limited in functionality for specific formats such as JSON and HTML.
Finally, we encounter also challenges related to engineering  parallelization and multiprocessing  of prompts, since the concurrent execution is non-trivial, especially with intricacies of Python process management.

\paragraph{Embedding Measure}
Our empirical measure is limited by the expressiveness of the embedding model and how well it captures the nuances in similarities between two representations. 
Furthermore, the obtained similarity scores are highly non-linear and difficult to interpret.
For instance, two representations may address the same topic, such as the problem description and its respective solution, however, when measuring their similarity we obtain similarity scores of $\sim70\%$.
We normalize this by subtracting an inherent baseline and randomness effect, however, to ensure a more holistic and robust measurement we would need a significantly larger amount of baselines and experiments. Since we were very limited in the availability of development resources, and some presented models are only addressable through costly API walls. We are actively seeking sponsors to scale our solution and offer a more compelling benchmark suite in the future.

\paragraph{Model Capabilities}

An obvious limitation revolves around the fixed context window size of the underlying language models. 
Despite the expansion of the context window in newer models such as GPT-4, the finite context still restricts the amount of data that can be processed in a single pass. All information outside the context needs to be added through information retrieval approaches, which come with their own challenges and limitations \citep{Gao:23}. This leads to side effects, including hallucination, given the model does not contain the necessary information to answer the prompted instruction, which makes it difficult to maintain long-term statefulness for  complex reasoning tasks and computational graphs.

\paragraph{Error Handling}

The complexity of error handling when evaluating complex expressions through function compositionality, especially between multiple modalities and different solvers, is another notable challenge. 
While SymbolicAI introduces mechanisms for error analysis and automated correction, these approaches are not infallible. They are often limited by the quality and expressiveness of the models, and the model's capacity to understand deeply nested logical constructs. We also note that for our evaluations, we disabled any remedy protocol, such as truncating prompts or retry schema.

\paragraph{Generalization}

This research is also limited by current LLM's capacity for reasoning and generalization. Although progress has been made, models are still prone to hallucinations and reasoning errors, especially when dealing with abstract, novel, or highly complex problem statements \citep{Marcus:20}. Furthermore, our framework's reliance on the model's ability to grasp the semantics of operations can be influenced by the training data and the model's innate biases and conceptual understanding \citep{Mahowald:23}.
We also point out that the initial development of SymbolicAI started with the GPT family of models, and we may encounter innate biases in prompt design and expressiveness when utilizing other reference models.
However, we also point out that prompt engineering instruction-based statements is not a reliable direction for improvement. We instead advocate for enhancing the resilience of models through fault tolerance, focusing on their ability to better follow semantic instructions, not syntactic idiosyncrasies.
Another concern is how to assess the disentanglement of evaluations of models on downstream tasks, to avoid testing on training samples, especially for closed-source solutions like GPT.

\paragraph{Interpretability and Transparency}

Finally, the issue of explainability and transparency in AI systems remains challenging. 
While SymbolicAI makes steps towards making computational processes more explicit and explainable through symbolic manipulations, understanding the internal logic and decision-making of LLMs remains an open problem.
This can hinder trust and adoption in sensitive applications where interpretability of predictions is important.

\subsection{Future Work}
The goal for Algorithm \ref{algo:evaluation_protocol} is to be utilized by an advanced learning agent. This agent, employing reinforcement learning methodologies \citep{Ouyang:22, Li:23, Rafailov:23}, could leverage our evaluation measure in existing benchmarks \citep{Milani:20, Swazinna:22, Schweighofer:22} as a means to obtain reward signals to addresses a central problem in reinforcement learning, namely credit assignment \citep{Sutton:84, Arjona:19, Holzleitner:20, Patil:20, Widrich:21, Dinu:22}. Over time, it aims to develop the ability to autonomously generate its own plans, efficiently schedule tasks and subtasks, and carefully select the most suitable tools for each task. Our protocol lays the groundwork for this agent to learn and expand its base set of capabilities \citep{Amaro:23}, moving towards more sophisticated, self-referential orchestration of multi-step tasks. We've already noticed that research is shifting towards this type of methodology \citep{Yuan:24}. 
Furthermore, in Section \ref{hierarchical_computation_graphs} we've only considered a sequential scheduler. 
However, our objective is to ultimately assess a non-sequential task execution model, allowing for dynamic insertion and out-of-sequence task execution. 
In addition, we are interested into exploring similarities of our work with \emph{Generative Flow Networks} \citep{Bengio:21, Bengio:21c, Lahlou:23}.
Lastly, we also discuss limitations in Appendix Section~\ref{sec:limitaiton} with further opportunities for future improvements.

\section{Conclusion}

In this work, we introduced SymbolicAI, a framework that unifies generative models with an array of solvers, blending the strengths of symbolic and sub-symbolic AI paradigms within a cohesive NeSy framework. 
SymbolicAI equips researchers and practitioners with a comprehensive toolkit to develop contextualized and explainable NeSy AI systems capable of addressing diverse challenges effectively. 
We also introduce a quality measure and a benchmark for comparing and evaluating a wide range of computational tasks.
SymbolicAI provides a basis for further research in advanced program synthesis, hierarchical computational graphs, the development of self-referential systems, and the integration of probabilistic methods with AI design for creating autonomous agents.

\section*{Acknowledgement}

The ELLIS Unit Linz, the LIT AI Lab, the Institute for Machine Learning, are supported by the Federal State Upper Austria. We thank the projects Medical Cognitive Computing Center (MC3), INCONTROL-RL (FFG-881064), PRIMAL (FFG-873979), S3AI (FFG-872172), DL for GranularFlow (FFG-871302), EPILEPSIA (FFG-892171), AIRI FG 9-N (FWF-36284, FWF-36235), AI4GreenHeatingGrids (FFG- 899943), INTEGRATE (FFG-892418), ELISE (H2020-ICT-2019-3 ID: 951847), Stars4Waters (HORIZON-CL6-2021-CLIMATE-01-01). We thank Audi.JKU Deep Learning Center, TGW LOGISTICS GROUP GMBH, Silicon Austria Labs (SAL), FILL Gesellschaft mbH, Anyline GmbH, Google, ZF Friedrichshafen AG, Robert Bosch GmbH, UCB Biopharma SRL, Merck Healthcare KGaA, Verbund AG, GLS (Univ. Waterloo), Software Competence Center Hagenberg GmbH, Borealis AG, T\"{U}V Austria, Frauscher Sensonic, TRUMPF, the NVIDIA Corporation and Atlas.

We extend our appreciation to Andreas Windisch and Clemens Wasner of AI Austria for their unwavering support. Their valuable feedback, connections, and facilitation of introductions within their expansive network have been instrumental to the progress of ExtensityAI.

Our gratitude also goes to Sergei Pereverzyev, whose enlightened guidance and thoughtful ideas have been a beacon for our research endeavors. Our thanks are equally extended to Gary Marcus, whose stimulating discussions sparked numerous innovative ideas incorporated into our framework.

We are equally grateful to Markus Hofmarcher, a friend and colleague whose informed counsel and stimulating discussions have significantly sharpened various facets of our study. Additionally, our thanks are due to Fabian Paischer and Kajetan Schweighofer, whose preliminary work and assistance have been of enormous benefit.

We are also grateful to our friends John Chong Min Tan and Tim Scarfe, whose communities have been a hub for exhilarating discussions. 
Their online presence and engagement have enriched the AI research landscape and broadened our perspectives.

Moreover, we wish to honor the memories of the cherished family members we lost in 2023. Their influence in our lives extended beyond personal bonds, and the principles they instilled in us continue to shape our journey. It is with great respect and affection that we acknowledge the indelible impact they have made, enabling us to persist in our scientific pursuits with determination and integrity.

\bibliography{references}
\bibliographystyle{iclr2024_conference}

\newpage
\appendix

\section{Broad AI and Neuro-Symbolic Systems}
\label{sec:broad_ai}

\begin{figure}[h!]
    \centering
    \includegraphics[width=1.0\linewidth]{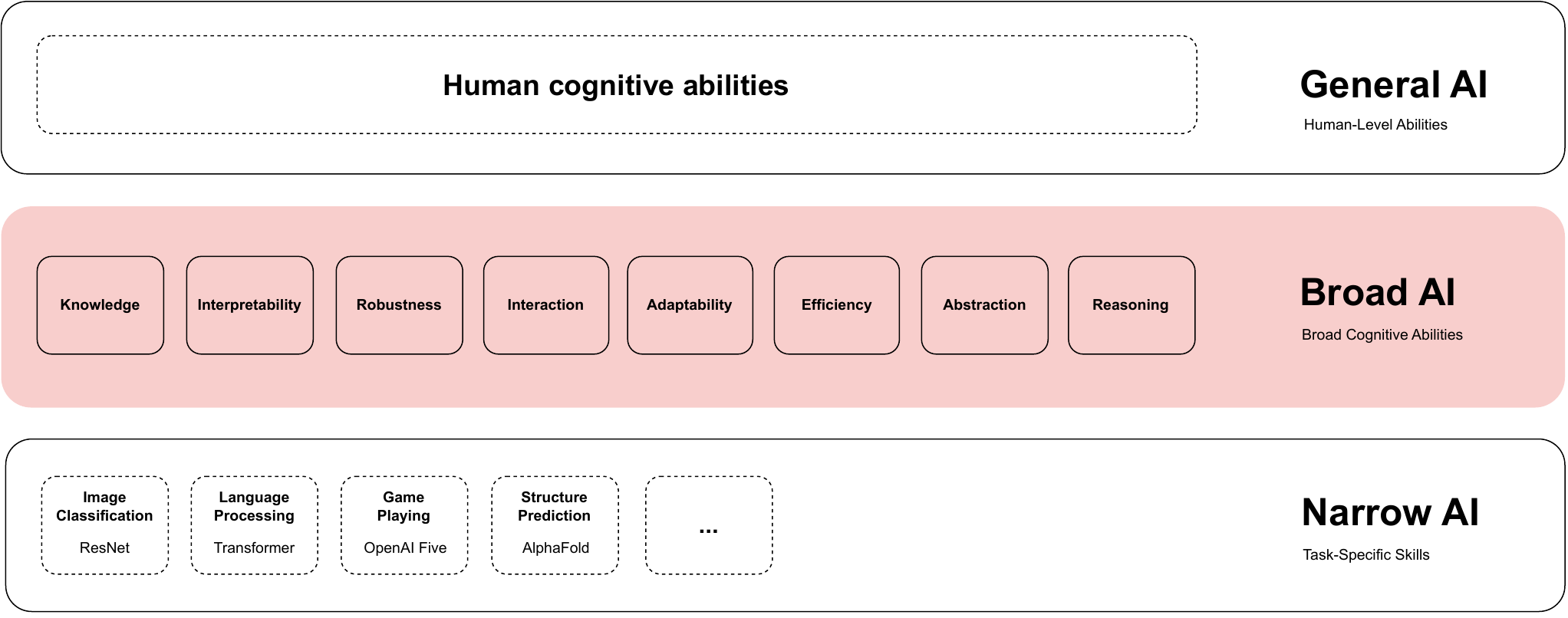}
    \caption{
    Hierarchical model of "\emph{cognitive}" abilities of AI systems \citep{Chollet:19, Hochreiter:22}. The figure contrasts the emergent paradigm of \emph{Broad AI} with current \emph{Narrow~AI} systems, showcasing Broad AI's wider range of capabilities, such as knowledge transfer, interaction, adaptability, robustness, abstraction, advanced reasoning, and efficiency. Broad AI aims to mimic human cognitive adaptability and robustness through advanced methodologies like few-shot learning, self-supervised contrastive learning, and context-sensitive sensory processing. Notably, Broad AI applies principles such as conceptual short-term memory and modern Hopfield networks \citep{Ramsauer:20} to better integrate context and memory, thus avoiding pitfalls like explaining away and short-cut learning. We acknowledge the potential of NeSy systems as a significant step towards AI systems capable of performing any cognitive task with human-like proficiency.}
    \label{fig:broad_ai}
\end{figure}

Our work focuses on broad AI \citep{Hochreiter:22} (see Figure~\ref{fig:broad_ai}) through the integration of symbolic and sub-symbolic AI methodologies. Broad AI extends beyond restricted focus on single-task performance of narrow AI. In broad AI, systems are engineered to handle a range of tasks with a high degree of autonomy, utilizing sensory input, accumulated experiences, and previously developed skills.

NeSy methods form the basis for developing new cognitive architectures \citep{Newell:56, Newell:57, Newell:72, Newell:90, Langley:09, Laird:22, Dawid:23, Sumers:23, LeCun:22a, Assran:23}.
This hybridization produces computational graphs capable of context-aware learning and reasoning, allowing AI to execute complex tasks with human-like flexibility. 

\begin{figure}[h!]
    \centering
    \includegraphics[width=0.8\linewidth]{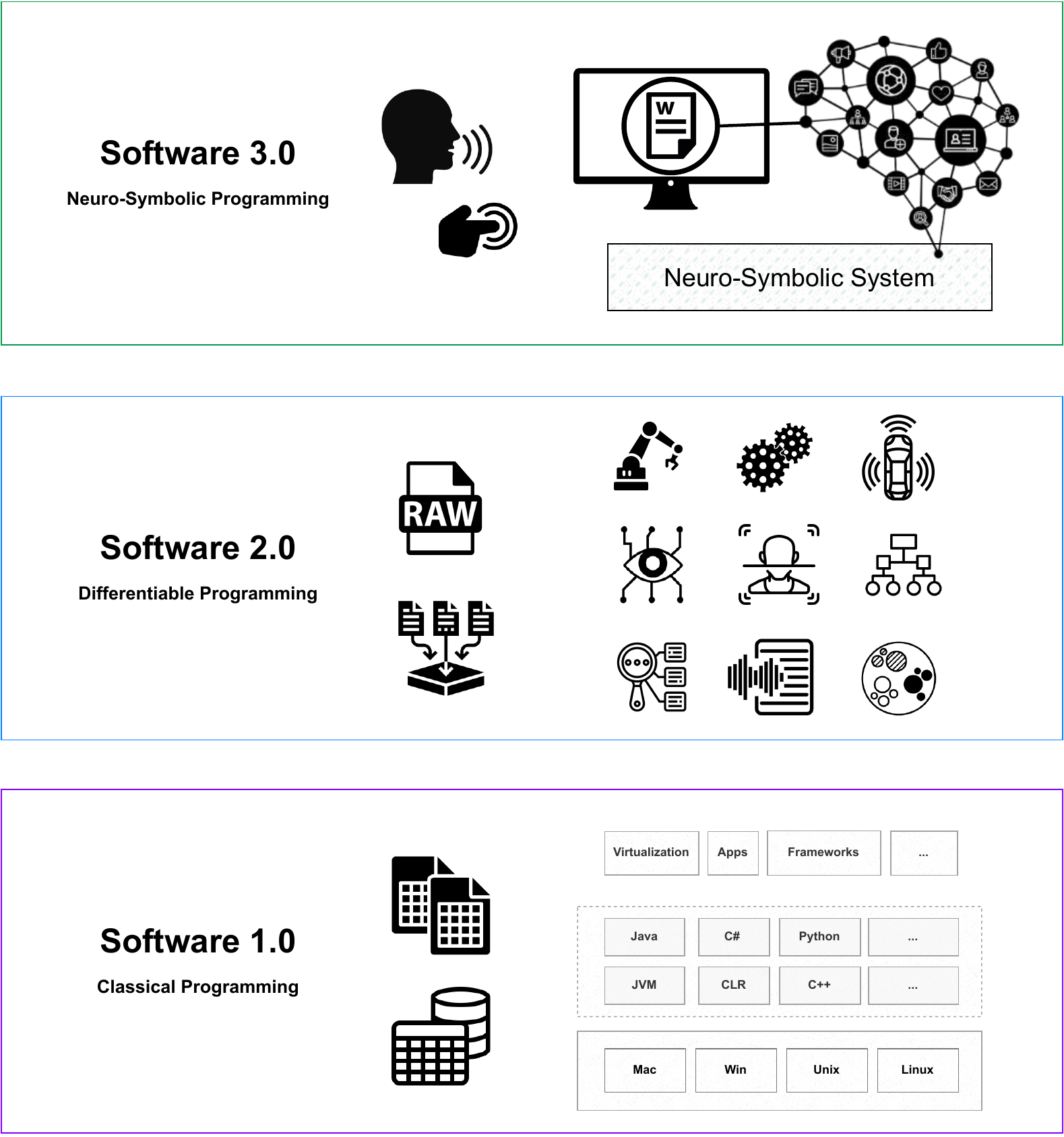}
    \caption{Evolution of software paradigms: From Software~1.0's rigid specification in classical programming to Software~2.0's data-driven and objective function-focused differentiable programming, leading to Software~3.0's NeSy systems that emphasize human-centric, interaction-based programming with computational graphs. This progression represents a shift from explicit task-specific programming to abstract, adaptive systems that cater to dynamic user preferences.}
    \label{fig:neuro_symbolic_interface_sw3}
\end{figure}

Borrowing nomenclature from \cite{Karpathy:17, Dilhara:21}, we refer to the next generation of software as Software~3.0, which consists of applications that are neither pre-determined at design time, nor learned through statistical inference, but triggered by an interaction which stimulates the realization of a computational graph analogous to \emph{neuromorphic circuits} \citep{Indiveri:11}, however,  purely established at inference time in the "\emph{thought}" process of a NeSy system.

\begin{figure}[h!]
    \centering
    \includegraphics[width=1.0\linewidth]{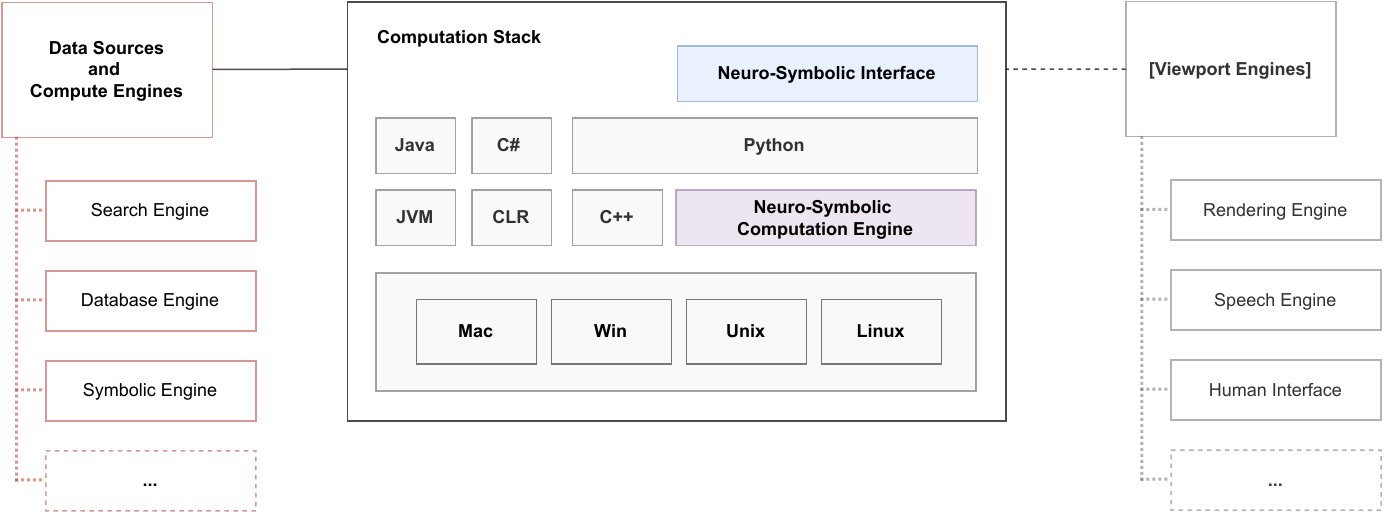}
    \caption{The illustration demonstrates the integration of Neuro-Symbolic computation within the contemporary programming stack. Probabilistic programming paradigms are embedded natively alongside traditional languages and environments, facilitated by interfaces to various data sources, compute engines, and human interaction tools, streamlining their adoption in practical and research applications.}
    \label{fig:ns_computation_stack}
\end{figure}

To enable such systems, we require a more native integration (see illustration in Figure~\ref{fig:ns_computation_stack}) of probabilistic programming paradigms into our contemporary programming stack, and make their utilization a commodity for practitioners and researchers alike.

\subsection{Broader Impact}

With LLMs becoming more and more accessible, progress recently made possible by the vast open source contributions from \cite{Koepf:23, Touvron:23, alpaca, xu2023baize, koala_blogpost_2023, biderman2023pythia}, embedded accelerators for LLMs --- or more generally NeSY engines --- will be ubiquitous in future computation platforms, such as wearables, smartphones, tablets, consoles, or notebooks. 
Although current execution cycles are slow and error-prone, we expect to see further performance gains through improved operating system level optimizations, dedicated GPU-centric hardware refinement, and improved software interoperability. 
We believe that modern programming paradigms should natively support probabilistic concepts and provide a boilerplate-free set of features for constructing and evaluating generative computational graphs. 
This includes but is not limited to compositional, parallelizable, and simulation-based executions with polymorphic operations and self-referential structures. 
Current programming languages often have disjointed or makeshift solutions for these concepts in the context of generative processes. 
We believe integral probabilistic support for these concepts into modern software and hardware will unlock new programming paradigms that can fully take advantage of generative architectures. 
We hope the community will consider these ideas as essential components of contemporary computing.

We also expect to see significant progress by processing central language concepts through system-on-a-chip (SoC) solutions of pre-trained models, with linear probing layers for hot-swappable weight exchange of task-specific projections and executions.
A wide range of functionalities can be then offloaded to probabilistic programming languages to operate on dedicated symbols and streamline the vector-valued mappings between the concept space and underlying problem space, avoiding defining boilerplate code to load and unload network weights. 

Furthermore, we believe that many gains in representational stability and consistency may be obtained through multi-modal data training and improved alignment based on operator learning oriented functionalities and workflow-related scoring functionalities, analogous to our introduced quality measure. 
Gains in representational stability also benefit self-instruction and self-referential sub-process evaluations, which enable the dynamic creation and evaluation of complex hierarchical computational graphs.
This will enable online learning models to perform, in real-time, skill acquisition of complex concepts with only one or few examples at inference time.
We believe this will enable the creation of autonomously self-evolving cognitive architectures \citep{Langley:09, Dawid:23, Sumers:23}.
We therefore see an inherent connection to generative design as an analogy for creating coherent and stable "\emph{thought}" computational graphs, and believe this paves the path toward broad AI systems (see Section~\ref{sec:broad_ai}) and is a requirement for developing artificial general intelligent agents.

Finally, we also wish to express our concern about recent economic trends in the deep-tech industry, where we observe AI-related concentration of data and resources, coupled with a tendency towards closed-source practices. 
We strongly advocate for increased transparency and exchange of ideas to ensure a diverse and collective growth in our socio-economic landscape. 
Therefore, we push towards a democratic and open-source initiative. 

\section{Connection between Fréchet Distance and Maximum Mean Discrepancy}
\label{sup:mmd_fd}

Let us consider a Gaussian kernel defined by the expression
\begin{equation}
K(x,y) = \exp\left(-\frac{\|x - y\|^2}{2\sigma^2}\right),
\end{equation}
where $\sigma$ is the bandwidth parameter of the kernel and $\|\cdot\|$ denotes the Euclidean norm. 
Utilizing $K$, we can now construct a measure of distance between distributions, by embedding them into the Reproducing Kernel Hilbert Space (RKHS) induced by $K$, using kernel mean embeddings. The resulting distance is called the Maximum Mean Discrepancy (MMD).

More precisely, the MMD between two probability distributions $P$ and $Q$ is encoded in the RKHS through mean embeddings, which can be expressed as
\begin{equation}
\text{MMD}^2(P,Q) = \left\|\mathbb{E}_{x \sim P}[\phi(x)] - \mathbb{E}_{y \sim Q}[\phi(y)]\right\|^2_{\text{RKHS}},
\end{equation}
where $\phi(\cdot)$ represents the feature mapping to the RKHS corresponding to the kernel $K$.

On the other hand, for multivariate Gaussian distributions, we can use the Fréchet distance as a measure of similarity, which is nothing but the associated Wasserstein-2 distance, for which an explicit formula is available in the Gaussian case. The resulting expression is as follows \citep{Dowson:82}:
\begin{equation}
d^2(X_1,X_2) = \|\mu_1 - \mu_2\|^2_2 + \text{Tr}\left(C_1 + C_2 - 2\left(C_1C_2\right)^{\frac{1}{2}} \right),
\end{equation}
where $X_1 \sim \mathcal{N}(\mu_1, C_1)$ and $X_2 \sim \mathcal{N}(\mu_2, C_2)$, and $\text{Tr}(\cdot)$ indicates the trace of a matrix.

To establish an approximation of the Fréchet distance using the Gaussian kernel, we take $C_1 = \sigma^2 I$ and $C_2 = \sigma^2 I$ as identity covariance matrices scaled by $\sigma^2$. This assumption allows us to focus solely on the disparities in mean vectors:
\begin{equation}
d^2(X_1, X_2) \approx \|\mu_1 - \mu_2\|^2_2,
\end{equation}
setting aside the effect of different covariance structures.

Given these conditions, we attempt to argue that the Fréchet distance behaves similarly as MMD:
\begin{equation}
d^2(X_1,X_2) \approx \|\mu_1 - \mu_2\|^2_2 \approx \text{MMD}^2(P,Q),
\end{equation}
Heuristically, at least for small $\|\mu_1 - \mu_2\|$, also the associated kernel evaluations $K(X_1,X_2)$ tend to be small (see also \cite{Hochreiter:97nc1}), which leads to a small MMD, if we ignore the terms associated to $K(X_1,X_1)$, $K(X_2,X_2)$ (which cancel out due to same covariance structure).

In the next section, we want to further elaborate on the MMD and a possible score, that can be derived from it.
\subsection{Extended Simplification of the MMD Calculation}

To understand the simplification of the MMD, we are formally expressing the MMD in terms of kernel sums over pairs of samples within and across two samples $X$ and $Y$:
\begin{equation}
\text{MMD}^2(X, Y) = \frac{1}{m(m-1)} \sum_{i} \sum_{j \neq i} k(x_i, x_j) - \frac{2}{mn} \sum_{i=1}^m \sum_{j=1}^n k(x_i, y_j) + \frac{1}{n(n-1)} \sum_{i} \sum_{j \neq i} k(y_i, y_j),
\end{equation}
where $m$ and $n$ are the sizes of samples $X$ and $Y$, respectively.

Empirical observations have led to the conclusion that the within-sample terms $\sum_{i} \sum_{j \neq i} k(x_i, x_j)$ and $\sum_{i} \sum_{j \neq i} k(y_i, y_j)$ cancel out the cross terms $\sum_{i=1}^m \sum_{j=1}^n k(x_i, y_j)$ under certain conditions. This can be due to the following:

\begin{itemize}
\item In high-dimensional embedding spaces, distributions of embedding vectors are often closely related and normally distributed.
\item If the samples $X$ and $Y$ are drawn from distributions $P$ and $Q$ where their mean embeddings are nearly orthogonal in the RKHS, it is the dissimilarity across samples, rather than that within, that is most relevant.
\end{itemize}

Therefore, under these specific conditions, it becomes justifiable to focus on the cross-terms, yielding the following proposal for a distance measure:
\begin{equation}
\widetilde{\text{MMD}^2}(X, Y) \approx \frac{2}{mn} \sum_{i=1}^m \sum_{j=1}^n k(x_i, y_j).
\end{equation}

\section{Structure}
\label{sup:framework_structure}

\paragraph{Primitives}
In the SymbolicAI framework, at the core lies the concept of Primitives and the dynamic type creation of \texttt{Symbol} objects, which are central to inherit types of behaviors. 
Primitives are pre-defined operations that act on \texttt{Symbol} objects, encapsulating basic operations, such as arithmetic, logic, or casting operations, to name a few. 
These operations are crucial to the framework's versatility and form the foundation for more complex interactions within computational graphs.
Essentially, they can be viewed as contextualized functions that accept a \texttt{Symbol} object, send it to the NeSy engine for evaluation, and return one or more new objects, primarily new symbols.
One of the key features of operations is their polymorphism, which allows for them to be applied to various data types, such as strings, integers, floats, lists, and more, with different behaviors depending on the specific object instance.
To execute operations, we utilize the \texttt{Symbol} object's \texttt{value} attribute containing the original data type.
This will be then sent as a string representation to the engines to execute the needed operations. 
Consequently, all values are cast to a string representation. 
Remember, this was our implicit assumption (see Section~\ref{sec:symandexpr}).
For custom objects, it is essential to define a suitable \texttt{\_\_str\_\_} method to cast the object to a string representation while preserving the object's semantics.

\paragraph{Symbol Objects Creation and Dynamic Typing}

A \texttt{Symbol} object is a versatile entity that can encapsulate a variety of data types and behaviors. The creation of \texttt{Symbol} objects is facilitated through a metaclass, which enables the dynamic typing of these objects to inherit behaviors from a collection of primitives. 
This dynamic typing system is important for extending the functionality of \texttt{Symbol} objects beyond simple data containers; they contain specific behaviors appropriate for the operations they will perform. 
For instance, a \texttt{Symbol} object may possess the behaviors of arithmetic computations, string manipulations, or even logical comparisons, depending on the defined primitives.

\paragraph{Type Inheritance and Expression Creation}

\begin{figure}[h!]
    \centering
    \includegraphics[width=1.0\linewidth]{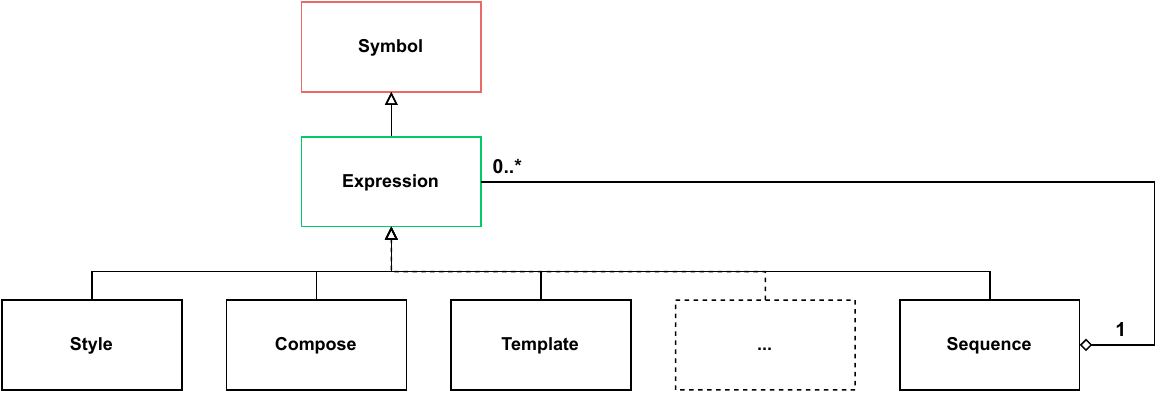}
    \caption{
    Class diagram showing the inheritance and composition relationships among \texttt{Symbol}, \texttt{Expression}, and other inherited classes.
    \texttt{Symbol} serves as a base class for \texttt{Expression} where all the other types are derived from. Other types may contain or associate with zero or more \texttt{Symbol} types. 
    For example, we illustrate how the \texttt{Sequence} derives from \text{Expression} and the multiplicity \texttt{'0..*'} indicates that a \texttt{Sequence} can contain any number of \texttt{Expression} instances.}
    \label{fig:symbol_inheritance}
\end{figure}

Type inheritance in SymbolicAI is leveraged to create new expressions, which are specialized forms of \texttt{Symbol} objects designed to represent parts of a computational graph.
Expressions extend the capabilities of \texttt{Symbol} objects by providing a structured way to define complex functionalities that can later be evaluated to produce new \texttt{Symbol} objects or modify existing ones. 
In SymbolicAI, an \texttt{Expression} object inherits the properties of \texttt{Symbol} objects while also being able to define its own unique behavior through a \texttt{forward} method, which is analogous to a computational graph node's evaluation function. Figure~\ref{fig:symbol_inheritance} gives an overview of an exemplary inheritance branch.
Each \texttt{Expression} must feature a \texttt{forward} method, which must be overwritten to define its behavior. 
The inherited \texttt{\_\_call\_\_} method invokes the \texttt{forward} method, evaluating the expression and returning the result. 
This design pattern facilitates lazy evaluation of expressions, allowing for complex composition of expressions.

Inherited from the \texttt{Symbol} class, the \texttt{\_sym\_return\_type} and \texttt{static\_context} properties establish the context in which the current \texttt{Expression} operates. 
The \texttt{static\_context} impacts all operations of the current \texttt{Expression} subclass, while the \texttt{\_sym\_return\_type} guarantees the acquisition of the desired return object type post-evaluation. 
Typically, this returns the current type but can be configured to return a different type.
A more in-depth examination of both notions will be provided in the following section.

\paragraph{Utilizing Decorators for Operation Definition}

Decorators serve as a bridge between the declarative nature of symbolic operations and the imperative execution model of programming languages. By augmenting function definitions with decorators, the framework can dynamically assign operations to \texttt{Symbol} or \texttt{Expression} objects, which are then interpreted by the underlying NeSy engine or traditional solvers.

For example, the \texttt{@core.logic} decorator can be used to augment a \texttt{Symbol} object with the capability to perform logical \texttt{and}, \texttt{or}, or \texttt{not} operations contextually. 
Similarly, the \texttt{@core.combine} decorator allows the framework to define the semantics of combining or adding two symbolic values, regardless of their underlying data representations.

\begin{lstlisting}[language=Python]
# Example of using decorators to define logical operations
@core.logic(operator='and')
def _some_logic(self, other):
  # implementation logic here
  pass
\end{lstlisting}

\paragraph{Aspect-Oriented Programming}

The aspect-oriented programming paradigm offers a functional approach for extending or modifying the behavior of functions or methods without altering their code directly. 
This adheres to the principles of modularity and separation of concerns, as it allows for the isolation of specific functionalities while maintaining the original function's core purpose. 
By wrapping the original function, decorators provide an efficient and reusable way of adding or modifying behaviors. 
For instance, SymbolicAI integrates the zero- and few-shot learning with default fallback functionalities of pre-existing code. 

Decorators brings several advantages \citep{beazley2009python, martelli2005python, summerfield2010python, lutz2013python}:

\begin{itemize}
    \item {\bf Reusability:} 
    Decorators promote code modularity, enhancing code reusability and contributing to software maintainability. 
    This advantage is particularly salient when managing a variety of operations, reducing redundancy and simplifying the integration of new functionalities.
    \item {\bf Composition:} 
    Decorators support function composition, allowing developers to construct complex functionalities from pre-existing code blocks without the need to expand the codebase or rely on complex inheritance hierarchies.
    \item {\bf Adaptability:} 
    Through decorators we can easily modify or extend the behavior of operations without changing their core implementation. 
    This flexibility facilitates the generation of adaptive workflows and reliable fallback mechanisms when experimental implementations do not fulfill required constraints. 
\end{itemize}

\paragraph{Symbol Class and Computational Graph Elements}
A computational graph in the SymbolicAI framework is an assembly of interconnected \texttt{Symbol} objects, each encapsulating a unit of data and the operations that can be performed on it. 
The exchange between these symbols forms a highly modular and interpretable system, capable of representing complex workflows.

The \texttt{Symbol} class is an abstraction representing data and context. 
It holds not only the value itself, but metadata that guides its transformation and interpretation. 
Through inheritance and compositionality, the \texttt{Symbol} can be extended into more complex expressions, and becoming nodes in a computational graph. 
Each \texttt{Symbol} instance can optionally contain a reference to its parent and children, naturally forming a directed graph structure where the nodes are symbols and edges represent relationships between a symbol and its derivative computations.

The \texttt{Linker} class, is a metadata subclass, and tracks relationships and results, effectively annotating the graph with execution details. 
It keeps records of nodes' keys, allowing quick retrieval of related computational outcomes within the graph, and aids in tasks such as error tracing and debugging. 

A central concept in this structure is the notion of \texttt{root}, which points to the origin of the computational sequence. 
Accessing the root allows backtracking through the graph, making it possible to aggregate results and inspect the flow of computation that led to the current node.

The computational graph's structure is further enriched by properties like \texttt{nodes}, \texttt{edges}, and \texttt{graph} itself, which collectively enable the comprehensive query of the computation's topology.
These properties are used internally to enable graph visualizations, which are useful for debugging and analysis.

\paragraph{Expression of a Computational Graph}
In practice, consider the \texttt{Expression} class, which extends the functionality of the \ttt{Symbol} class. 
When composing a \texttt{Sequence} of \texttt{Expression} objects, we are effectively composing operations in a predetermined order.

For instance, an expression like:

\begin{lstlisting}[language=Python]
Sequence(
  Clean(),
  Translate(),
  Outline(),
  Compose(),
)
\end{lstlisting}

represents a procedure that first cleans data, then translates it, outlines the essential information, and composes it into a finalized form.
When this sequence is executed, the operations unfold in the exact order specified, with each step receiving the output of its predecessor as input and passing its result to the successor.


\section{Installation}
\label{sec:installation}

The installation of the SymbolicAI framework is straightforward and can be done through the Python package manager \texttt{}{pip}. To install SymbolicAI, open a terminal and execute the following command in your current python environment:

\begin{lstlisting}[language=Bash]
pip install symbolicai
\end{lstlisting}

This command will install the latest version of SymbolicAI along with its core dependencies, enabling the integration of the framework into Python applications. If you intend to utilize the framework with local engines\footnote{\ The local engines are experimental and are run on your local machine. For more details, refer to the "Local Neuro-Symbolic Engine" section in the documentation.}, or with engines powered by external APIs such as OpenAI's API, additional installation steps are required.

\subsection{Engine Configuration}

Before the first run, it is necessary to configure the required modules and optionally set necessary API keys to activate the respective engines.
This can be done in multiple ways, but we recommend doing it through the configuration wizard by running this command in the terminal:

\begin{lstlisting}[language=Bash]
symwzd
\end{lstlisting}

This step is essential to register the engines internally for subsequent runs. 

For instance, SymbolicAI includes OpenAI's GPT models as NeSy engine. 
To only set or change OpenAI API keys, the following command is used before starting a SymbolicAI instance:

\begin{lstlisting}[language=Bash]
# Linux / MacOS
export OPENAI_API_KEY="<OPENAI_API_KEY>"
\end{lstlisting}

After setting up the API keys, the SymbolicAI library is imported in Python as following:

\begin{lstlisting}[language=Python]
import symai
\end{lstlisting}

For more low-level changes, we store everything under the \texttt{\$HOME/.symai} folder, such as the \texttt{symai.config.json}, which stores every key, both registered and not registered.

\subsection{Optional Installations}

\begin{figure}[h!]
    \centering
    \includegraphics[width=1.0\linewidth]{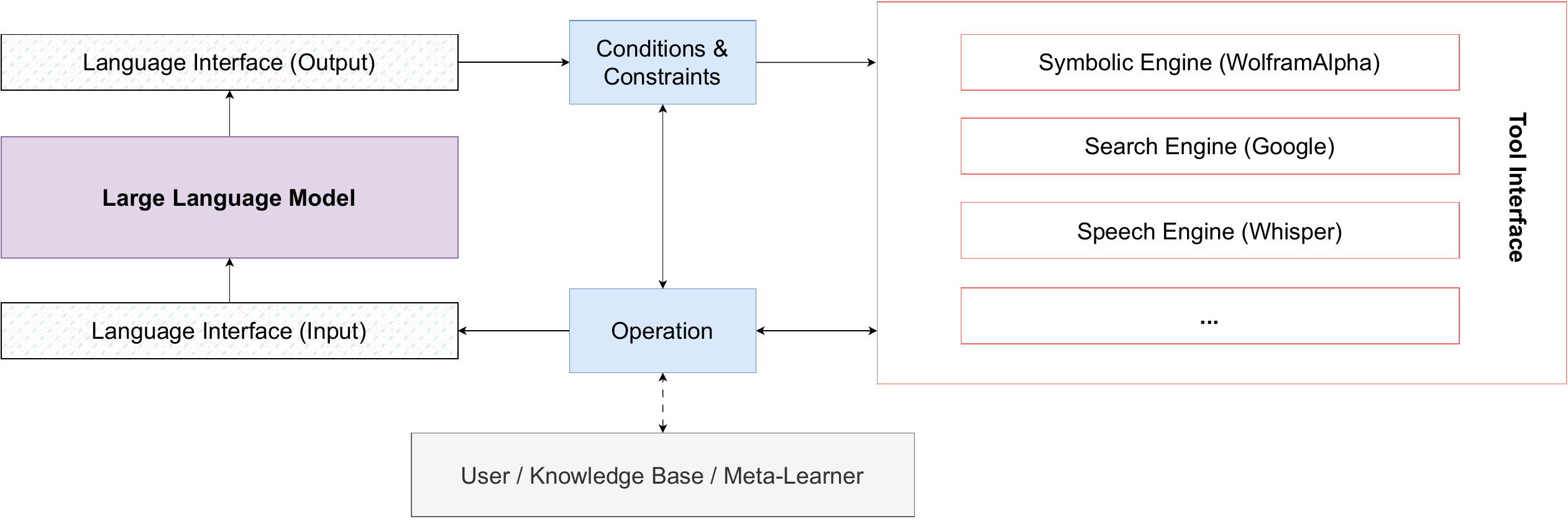}
    \caption{The SymbolicAI framework integrates a Large Language Model (LLM) with diverse tools and solvers through a conceptual interaction stack. The framework enables operations initiated by users, knowledge bases, or meta-learners to be processed by the LLM, which interfaces with specialized engines such as WolframAlpha and Whisper via conditions and constraints, enhancing the AI's problem-solving capabilities.}
    \label{fig:nsce_interaction_stack}
\end{figure}

The SymbolicAI framework is designed to leverage multiple engines for a variety of operations. 
To fully utilize these capabilities, you may install additional dependencies or set up the optional API keys for specific engines like WolframAlpha, SerpApi, and others. In Figure~\ref{fig:nsce_interaction_stack} we conceptually outline the connection between the utilization of an LLM and its interact with other tools and solvers. 
Instructions and operations can be initiated by any user, pre-scripted knowledge base or learned meta agent.

For instructions on additional installations, including the support of optional engines, refer to the documentation provided with the framework. 
This documentation will give detailed steps on installing optional dependencies and configuring additional API keys.

\section{Implementation Details}
\label{sup:impl_details}

Let us now define some \texttt{Symbol} objects and perform some basic manipulations. 

\subsection{Fuzzy Comparison}
For instance, let's consider fuzzy\footnote{\ Not related to fuzzy logic, which is a topic under active consideration.} comparisons.
Within SymbolicAI, it enables more adaptable and context-aware evaluations, accommodating the inherent uncertainties and variances often encountered in real-world data.

\begin{lstlisting}[language=Python]
import numpy

s = symai.Symbol('3.1415...')
s == numpy.pi
\end{lstlisting}

\begin{lstlisting}[language=Bash]
:[Output]: 
True
\end{lstlisting}

\subsection{Dynamic Casting}
By enabling sentence subtraction and dynamic casting within SymbolicAI, we utilize the generalization capability of NeSy engines to manipulate and refine text-based data, creating more meaningful and contextually relevant outcomes. 
The integration of dynamic casting with \texttt{Symbol} objects in our API allows the users to perform operations between \texttt{Symbol} objects and various data types, such as strings, integers, floats, lists, etc. without compromising on readability or simplicity. 

\begin{lstlisting}[language=Python]
s = symai.Symbol('Hello my enemy') 
s - 'enemy' + 'friend'
\end{lstlisting}

\begin{lstlisting}[language=Bash]
:[Output]: 
<class 'symai.expressions.Symbol'>(value=Hello my friend)
\end{lstlisting}

\subsection{Symbols and Embeddings}

It is worth noting that encoding a complex object into a string sometimes precludes the object reconstitution.
However, this concern does not substantially impede our methodology as we can employ approximations or proxy representations stored by the vector-valued property to effectively re-map objects.
These representations are obtained through respective embedding models.

\subsection{Translation}
In today's increasingly interconnected world, translation between languages is fundamental, making it an essential feature. 

\begin{lstlisting}[language=Python]
s = symai.Symbol("Welcome to our tutorial.")
s.translate('German')
\end{lstlisting}

\begin{lstlisting}[language=Bash]
:[Output]: 
<class 'symai.expressions.Symbol'>(value=Willkommen zu unserem Tutorial.)
\end{lstlisting}

\subsection{Filtering, Ranking, Extraction}
Incorporating data-agnostic operations like filtering, ranking, and pattern extraction into our API allow the users to easily manipulate and analyze diverse data sets.

\begin{lstlisting}[language=Python]
s = symai.Symbol(numpy.array([1, 2, 3, 4, 5, 6, 7]))
s.rank(measure='numerical', order='descending')
\end{lstlisting}

\begin{lstlisting}[language=Bash]
:[Output]: 
<class 'symai.expressions.Symbol'>(value=['7', '6', '5', '4', '3', '2', '1'])
\end{lstlisting}

\subsection{Implications}
One of the main objectives behind developing SymbolicAI was to facilitate reasoning capabilities in conjunction with the statistical inference inherent in LLMs. 
Consequently, we can carry out deductive reasoning operations through the \texttt{Symbol} objects. 
For instance, it is feasible to establish a series of operations with rules delineating the causal relationship between two symbols. 
The subsequent example illustrates the utilization of the \texttt{\&} operator to compute the logical implication derived from the interaction of two distinct symbols.

\begin{lstlisting}[language=Python]
s1 = symai.Symbol('The horn only sounds on Sundays.') 
s2 = symai.Symbol('I hear the horn.')
s1 & s2
\end{lstlisting}

\begin{lstlisting}[language=Bash]
:[Output]:
<class 'symai.expressions.Symbol'>(value=It is Sunday.)
\end{lstlisting}

In the above example, the \texttt{\&} operator overloads the \texttt{and} logical operator and extends its functionality. 
Furthermore, we can establish more sophisticated logical operators for \texttt{and}, \texttt{or}, and \texttt{xor} that can be grounded in formal proofs and utilize the NeSy engine to parse data structures before evaluating the expressions. 
This enables the definition of bespoke operations for executing intricate and robust logical operations, incorporating constraints to validate outcomes and guide the computation towards the desired behavior. 

\subsection{Custom operations}
The following example demonstrates how to define a custom \texttt{==} operation by overriding the \texttt{\_\_eq\_\_} method and providing a custom prompt object with a list of examples:

\begin{lstlisting}[language=Python]
import symai

class Demo(symai.Symbol):
  def __eq__(self, other) -> bool:
    # define nested function
    @symai.core.equals(examples=symai.Prompt([
      "1 == 'ONE' =>True",
      "'six' == 7 =>False",
      "'Acht' == 'eight' =>True",
      ...
    ]))
    def _func(_, other) -> bool: # [optional] cast return type           (1. below)
      return False             # [optional] default behavior on failure  (2. below)
    return _func(self, other)
\end{lstlisting}

As illustrated in the example, this is also the method we used to implement basic operations in \texttt{Symbol}, namely by defining local functions that are then decorated with the respective operation decorator from the \texttt{symai.core.py} file. 
The \texttt{symai.core.py} is a collection of pre-defined operation decorators that can be quickly applied to any function.
We use locally defined functions instead of directly decorating the main methods for two reasons: 
\begin{enumerate}
    \item We want to cast return types of the operation outcome to symbols or other derived classes thereof.
    \item We do not necessarily want all of our operations to be sent to the NeSy engine and might need to implement a default behavior.
\end{enumerate}
This is achieved through the \texttt{\_sym\_return\_type} method, which can provide contextualized behavior based on the defined return type.
More details can be found in the actual \texttt{Symbol} class.

In the context of LLMs, zero- and few-shot learning domains have emerged as essential techniques \citep{Yao:23React, shinn2023reflexion, kim2023language, Wei:22CoT, lyu2023faithful, pitis2023boosted, Madaan:23, Wang:22Instruct, Ye:23}\footnote{\ This is by no means an exhaustive list, we only point the reader to some very influential and recent research.} to enable models to generalize from limited training data and adapt to new tasks without requiring extensive retraining. 
This capability to learn and perform tasks with minimal examples is highly desirable, as it reduces the need for large labeled data sets and allows for faster deployment in new applications.
In this section, we demonstrate how our Symbolic API incorporates Python decorators to define custom operations in the zero- and few-shot domains.

Consider the following example, where we define a custom operation to generate a random integer between $0$ and $10$ using the Symbolic API and Python decorators:

\begin{lstlisting}[language=Python]
import symai

class Demo(symai.Symbol):
  def __init__(self, value = '') -> None:
    super().__init__(value)
    
  @symai.core.zero_shot(prompt="Generate a random integer between 0 and 10.",
                        constraints=[
                          lambda x: x >= 0,
                          lambda x: x <= 10
                        ])
  def get_random_int(self) -> int:
    pass
\end{lstlisting}

In this example, the \texttt{@symai.core.zero\_shot} decorator is used to define a custom operation that does not require any examples, as the prompt is expressive enough.
The \texttt{zero\_shot} decorator takes in two arguments: \texttt{prompt} and \texttt{constraints}. 
The prompt defines the conditioning for our desired operation behavior, while the constraints are used to validate the computed outcome, ensuring it meets our expectations.
If the constraints are not fulfilled, the implementation would resort to the specified default implementation or the default value.
If neither is provided, the Symbolic API raises a \texttt{ConstraintViolationException}.
The return type in the example is defined as \texttt{int}. 
The resulting value from the wrapped function must be of type \texttt{int} because of the specific implementation of the auto-casting realized through \texttt{->}. 
If the cast fails, the Symbolic API raises a \texttt{ValueError}. 
If no return type is specified, the return type defaults to \texttt{Any}.

The \texttt{@symai.core.few\_shot} decorator is a generalized version of \texttt{@symai.core.zero\_shot} and is used to define custom operations requiring examples. 
The function signature of the \texttt{few\_shot} decorator is as follows:

\begin{lstlisting}[language=Python]
def few_shot(prompt: str,
             examples: Prompt, 
             constraints: List[Callable] = [],
             default: Any = None, 
             limit: int = 1,
             pre_processor: Optional[List[PreProcessor]] = None,
             post_processor: Optional[List[PostProcessor]] = None,
             **wrp_kwargs):
\end{lstlisting}

The behavior of the \texttt{prompt} and \texttt{constraints} attributes is similar to the \texttt{zero\_shot} decorator.
The \texttt{examples} and \texttt{limit} arguments are new, with \texttt{examples} defining a list of instructions conditioning the NeSy engine, and \texttt{limit} specifying the maximum number of examples returned. 
The \texttt{pre\_processor} and \texttt{post\_processor} arguments accept lists of \texttt{PreProcessor} and \texttt{PostProcessor} objects, respectively, which are utilized to process the input before being fed into the NeSy engine and the output before being returned to the user.
The \texttt{wrp\_kwargs} argument passes additional arguments to the wrapped method, streamlining them towards the NeSy engine, or other engines.

\subsection{Prompting}

In this section, we discuss the design of prompts and their role in shaping the behavior of operations. 
Prompts serve as containers for information necessary to define specific operations, and the \texttt{Prompt} class serves as the base class for all the other \texttt{Prompt} classes.
Consider the following example, where we define a \texttt{Prompt} for comparing two values through the NeSy engine.
In it, when the \texttt{<=} operation on two \texttt{Symbol} objects will be resolved, the NeSy engine evaluates them in the context of the \texttt{CompareValues} prompt.

\begin{lstlisting}[language=Python]
class CompareValues(symai.Prompt):
  def __init__(self) -> symai.Prompt:
    super().__init__([
      "4 > 88 =>False",
      "-inf < 0 =>True",
      "inf > 0 =>True",
      "4 > 3 =>True",
      "1 < 'four' =>True",
      ...
    ])
\end{lstlisting}

Evaluating a fuzzy comparison statement:
\begin{lstlisting}[language=Python]
res = symai.Symbol(1) <= symai.Symbol('one')
\end{lstlisting}

Output of the evaluation:
\begin{lstlisting}[language=Bash]
:[Output]:
True
\end{lstlisting}

This evaluation returns \texttt{True}, as the fuzzy comparison operation conditions the engine to compare the two \texttt{Symbol} objects based on their semantic meaning. 
More generally, the semantics of \texttt{Symbol} operations may vary depending on the context hierarchy of the \texttt{Expression} class and the operations used. 
We used three main prompt designs: \emph{Context-based Prompts}, \emph{Operational Prompts}, and \emph{Templates}. 
Prompts can be curated either through inheritance or composition. 
For instance, the \emph{static context} can be defined by inheriting from the \texttt{Expression} class and overwriting the \texttt{static\_context} property, while an \texttt{Operation} and \texttt{Template} prompt can be created by providing a \texttt{PreProcessor} to modify the input data.

We will now provide a more detailed explanation for each prompt design:

\begin{enumerate}
    \item Context-based Prompts are considered optional and can be defined in a static manner, either by sub-classing the \texttt{Expression} class and overwriting the \texttt{static\_context} property, or at runtime by updating the \texttt{dynamic\_context} property or passing a \texttt{payload} kwargs to a method. 
    Below is an example of \texttt{payload} kwargs through the method signature:
    
    \begin{lstlisting}[language=Python]
# creating a query to ask if an issue was resolve or not
s = symai.Symbol("<some-community-conversation>")
q = s.query("Was the issue resolved?")
# write manual condition to check if the issue was resolved
if 'not resolved' in q:
  # do a new query but payload the previous query answer to the new query
  s.query("What was the resolution?", payload=q)
  ...
else:
  pass # all good
    \end{lstlisting}
    
    Regardless of how the context is set, the contextualized prompt defines the desired behavior of \texttt{Expression} operations. 
    For instance, if we want to operate in the context of a DSL without having to overwrite each base class method, we can utilize this approach\footnote{\ See more details in this \href{https://github.com/ExtensityAI/symbolicai/blob/main/notebooks/Queries.ipynb}{notebook}.}.
    \item Operational Prompts define the behavior of an atomic operation and are therefore mandatory to express the nature of such an operation. 
    For example, the \texttt{+} operation is used to add two \texttt{Symbol} objects together, and the \texttt{+} operation prompt explains its behavior. 
    The \texttt{examples} kwargs provide another optional structure that conditions the NeSy engine with a set of instructions.
    \item Template Prompts are optional and encapsulate the resulting prediction to enforce a specific format. 
    For example, to generate HTML tags, we can utilize a curated \texttt{<html>{{...}}</html>} template. 
    This template enforces the NeSy engine to begin the generation process already in the context of an HTML tags format and not produce irrelevant descriptions about its task.
\end{enumerate}

\subsection{Complex expressions}
We will now attempt to obtain logical answers based on questions of the kind:
\begin{itemize}
  \item A line parallel to $y = 4x + 6$ passes through $(5, 10)$. What is the $y$-coordinate of the intercept?
  \item Bob has two sons, John and Jay. Jay has one brother and father. The father has two sons. Jay's brother has a brother and a father. Who is Jay's brother?
  \item Is 1000 bigger than 1063.472?
\end{itemize}

To solve these tasks, we would initially employ a series of operations to identify the most suitable engine for handling the specific requirements. 
Subsequently, we would prepare the input tailored to the selected engine. 

\begin{lstlisting}[language=Python]
val = "<one of the examples above>"

# First define a class that inherits from the \texttt{Expression} class
class ComplexExpression(symai.Expression): 
  # write a method that returns the causal evaluation
  def causal_expression(self):
    pass 

# instantiate an object of the class
expr = ComplexExpression(val)
# set WolframAlpha as the main expression engine to use
expr.command(engines=['symbolic'], expression_engine='wolframalpha')
# evaluate the expression
res = expr.causal_expression()
\end{lstlisting}

The implementation of \texttt{causal\_expression} could in principle look like this:

\begin{lstlisting}[language=Python]
def causal_expression(self):
  if self.isinstanceof('mathematics'):
    # get the mathematical formula
    formula = self.extract('mathematical formula')
    # verify which problem type we have
    if formula.isinstanceof('linear function'):
      # prepare for wolframalpha
      question = self.extract('question sentence')
      req = question.extract('what is requested?')
      # get coordinate point / could also ask for other points
      x   = self.extract('coordinate point (.,.)')
      # concatenate to the question and formula
      query = formula | f', point x = {x}' | f', solve {req}'
      res = self.expression(query) # send prepared query to wolframalpha

    elif formula.isinstanceof('number comparison'):
      res = formula.expression() # send directly to wolframalpha

      ... # more cases

  elif self.isinstanceof('graph construction'):
    sentences = self / '.' # first split into sentences
    graph = {} # define graph
    for s in sentences:
      sym = symai.Symbol(s)
      relations = sym.extract(
        # extract and split by pipe
        'connected entities (e.g. A has three B => A | A: three B)') / '|'
      for r in relations:
        ... # add relations and populate graph; reading suggestion

  ... # more cases
  
  return res
\end{lstlisting}

The aforementioned example demonstrates the utilization of the \texttt{causal\_expression} method, which allows us to extract information that can be resolved either manually or through external solvers, say WolframAlpha. 
Initially, when utilizing the GPT-3 backend, we anticipated a significant engineering effort to develop such a complex expression, as the GPT-3 backend frequently struggled with accurate information extraction and comparison resolution. 
However, we remained confident in the field's progress, specifically with fine-tuned models like RLHF ChatGPT. 
We were delighted to witness these challenges being further tackled through the latest GPT-4 model \citep{openai23}.

Furthermore, it is worth highlighting that, given sufficient data, we could refine methods for information extraction or knowledge graph construction from natural language, enabling more intricate reasoning tasks, such as those previously mentioned. 
We also direct readers to recent publications on Text-to-Graph translations, especially the very influential CycleGT \citep{Guo:20}. 
This approach allows us to answer queries by simply traversing the graph and extracting the required information.

Lastly, recent research \citep{kiciman23, Ellis:23} has demonstrated that algorithms based on GPT-3.5 and GPT-4 establish new state-of-the-art accuracy on multiple causal benchmarks, while also exhibiting unique capabilities previously considered exclusive to humans, such as generating causal graphs and identifying background causal context from natural language. 
This points to the potential for LLMs to be used alongside existing causal methods as proxies for human domain knowledge, reducing human effort in setting up causal analyses and ultimately accelerating the widespread adoption of causal methods. 
Moreover, recent advances in LLMs have opened new frontiers for research, practice, and adoption of causal reasoning, transforming the way causal analysis is conducted and broadening the scope of applications for our framework.

One of the most prominent illustrations of this concept is exhibited by Word2Vec \citep{mikolov13}. 
Word2Vec generates dense representations of words by training a shallow neural network to predict a word based on its neighboring words within a text corpus. 
These resulting vectors are extensively utilized in various natural language processing applications, including sentiment analysis, text classification, and clustering.

Drawing parallels with Word2Vec, our objective is to execute \emph{contextualized} operations on different symbols.
However, the key distinction lies in the fact that we operate within the natural language domain, as opposed to a vector space. 
Consequently, this grants us the capability to conduct arithmetic on words, sentences, paragraphs, and the like, while simultaneously validating the outcomes in a human-readable format.

The following example, we illustrate the methodology for evaluating such an expression through a string representation:

\begin{lstlisting}[language=Python]
s = symai.Symbol('King - Man + Woman')
s.expression()
\end{lstlisting}

\begin{lstlisting}[language=Bash]
:[Output]:
<class 'symai.expressions.Symbol'>(value=Queen)
\end{lstlisting}

In contrast to the \texttt{Symbol} object, the \texttt{Expression} represents a non-terminal symbol. 
It allows for further evaluation and extends the \texttt{Symbol} class by overwriting the \text{\_\_call\_\_} method.
It serves as the foundation for all other expressions and possesses additional capabilities, namely to \texttt{fetch} data from URLs, \texttt{search} the internet, or \texttt{open} files. 
These operations are intentionally separated from \texttt{Symbol}, as they do not utilize the \texttt{value} attribute of the \texttt{Symbol} class.

\subsection{Composition}

\subsection{Sequences}
Sequences offer a multitude of advantages in the realm of \texttt{Expression} objects, as they facilitate the creation of more sophisticated structural configurations. 
By embodying the \texttt{Sequence} expression, multiple expressions can be effectively evaluated at runtime, thus enhancing the flexibility, modularity, and adaptability of the framework. 

\begin{lstlisting}[language=Python]
# first import all expressions
from symai.components import *
# define a sequence of expressions
Sequence(
  Clean(),
  Translate(),
  Outline(),
  Compose(),
)
\end{lstlisting}

\subsection{Streams}

As demonstrated earlier, creating contextualized prompts refines the behavior of operations in the NeSy engine. 
However, this also consumes a considerable portion of the available context size. 
Given a limited context size, this constraint may pose challenges.
Fortunately, the \texttt{Stream} processing expression offers a solution by opening a data stream and performing chunk-based operations on the input stream.
\texttt{Stream} expressions can encapsulate other expressions. 
For instance, chunks can be managed through a \texttt{Sequence} expression, which permits multiple compositional operations sequentially. 
The example below illustrates the definition of a \texttt{Stream} expression:

\begin{lstlisting}[language=Python]
Stream(
  Sequence(
    Clean(),
    Translate(),
    Outline(),
    Embed()
  )
)
\end{lstlisting}

In this case, a stream is opened and a \texttt{Sequence} expression is passed, which cleans, translates, outlines, and embeds the input. Internally, the stream operation estimates the available model context size and segments the lengthy input text into smaller chunks transmitted to the inner expression. 
The returned object type is a generator.

\begin{figure}[h!]
    \centering
    \includegraphics[width=0.8\linewidth]{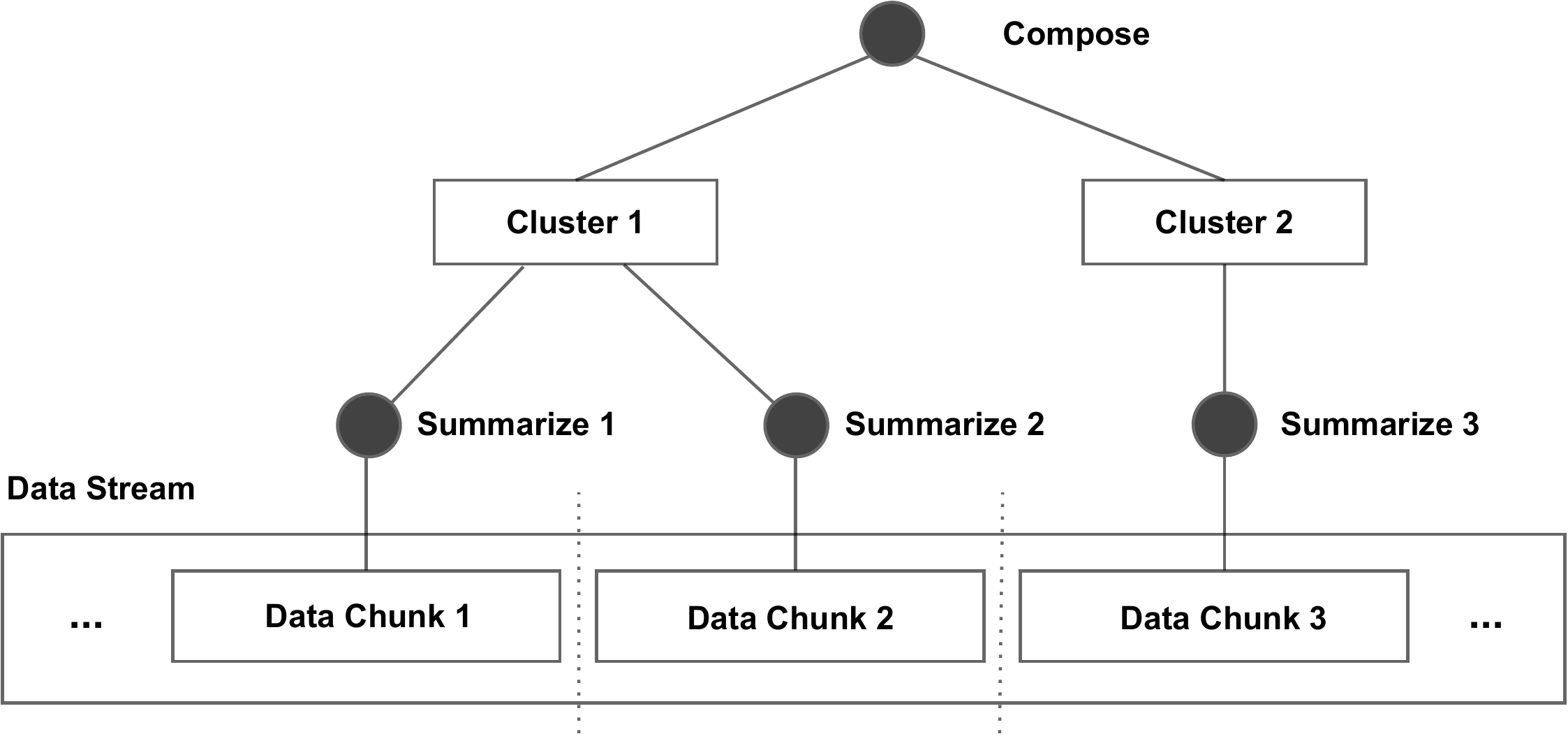}
    \caption{Stream processing expression in NeSy engine, illustrating data stream segmentation into chunks, each undergoing operations like cleaning, outlining, and embedding. The Cluster expression then merges chunks based on similarity, allowing contextually related information to be consolidated meaningfully. Node summaries are generated by extracting key labels from each cluster's content, overcoming context size limitations and maintaining shared information among processed chunks.}
    \label{fig:cluster_data_stream}
\end{figure}

The limitation of this approach is that the resulting chunks are processed independently, lacking shared context or information among them. 
To address this, the \texttt{Cluster} expression can be employed, merging the independent chunks based on their similarity, as it illustrated in Figure~\ref{fig:cluster_data_stream}.

By merging individual chunks by clustering their contents, contextually related information can be consolidated in a meaningful manner.
Additionally, the clustered information can be labeled by streaming through each cluster's content and extracting the most pertinent labels, yielding interpretable node summaries.

The complete example is depicted as follows:

\begin{lstlisting}[language=Python]
stream = Stream(
  Sequence(
    Clean(),
    Translate(),
    Outline(),
  )
)
        
s    = symai.Symbol('<some long text>')
res  = symai.Symbol(list(stream(s)))
expr = Cluster()
expr(res)
\end{lstlisting}

Subsequently, this process can be recursively repeated on each summary node to construct a hierarchical clustering structure.
As each node represents a summarized subset of the original information, the summary can function as an index. 
The resulting tree can be utilized to navigate and retrieve the original information, transforming the large data stream problem into a search problem.
Alternatively, vector-based similarity searches can be employed to identify similar nodes.
For searching within a vector space, dedicated libraries such as Annoy \citep{spotify:22}, Faiss \citep{johnson19}, or Milvus \citep{Wang:21Milvus} can be used.

In summary, \texttt{Stream} expressions offer the advantage of processing large data streams in a more efficient and organized manner, while also enabling the integration with other expressions like \texttt{Sequence} and \texttt{Cluster} expressions. 
These combinations allow for a more effective approach to handling context limitations, promoting the extraction of meaningful information and improving the overall performance of the framework.

\subsection{Error handling, debugging, and explainability}
Effective error handling and debugging are essential for ensuring the robustness and reliability of any software system, while explainability is essential for understanding the underlying behavior of the system, especially in the context of AI-driven applications. 
By developing a system that is both transparent and interpretable, we can gain valuable insights into the performance of the NeSy engines and identify potential areas for improvement.

\subsection{Error handling}
One of the fundamental aspects of the SymbolicAI API is being able to generate code. 
Consequently, errors may arise, and handling them contextually becomes vital.
In pursuit of a self-evolving API, we introduce the \texttt{Try} expression, which includes built-in fallback statements and automatically retries execution after performing dedicated error analysis and correction. 
This expression analyzes both the input and the error, conditioning itself to resolve the error by modifying the original code\footnote{\ This is similar to the recently released \href{https://github.com/Significant-Gravitas/Auto-GPT}{Auto-GPT} application.}. 
If the fallback expression succeeds, the result is returned; otherwise, the process is repeated for the specified number of retries. 
If the maximum number of retries is reached without resolving the issue, the error is raised again.

Consider the example of executing previously generated code that contains a syntax error. 
By employing the \texttt{Execute} expression, we can evaluate the generated code, which takes a symbol and proceeds with the execution. 
Despite the initial failure, the \texttt{Try} expression resolves the syntactic error, returning the corrected and evaluated code:

\begin{lstlisting}[language=Python]
expr = Try(expr=Execute())
s    = symai.Symbol('a = int("3,")') # some code with a syntax error
expr(s)
\end{lstlisting}

\begin{lstlisting}[language=Bash]
:Output:
a = 3
\end{lstlisting}

While not all errors can be resolved as easily as the demonstrated syntax error, we continue to explore more sophisticated error handling mechanisms, including streams and clustering to address errors in a hierarchical and contextual manner. 

\section{Evaluation Details}\label{sup:eval_details}
In a computational graph, the VERTEX score compares the distribution of the generated model answer at each node against a reference distribution by sampling multiple valid trajectories at each node for the reference distribution and accounting for randomness through some predefined random trajectories. 
For instance, one of the predefined random trajectories in our benchmark was the string of ASCII characters which are considered printable, namely {\texttt{0123456789abcdefghijklmnopqrstuvwxyzABCDEFGHIJKLMNOPQRSTUVWXYZ!"\#\$\%\&'()*+, -./:;<=>?@[]\^\_`\{|\}\~}}.
Moreover, the VERTEX score is particularly well suited for the evaluation of multi-step workflows and in contexts where the solution space is or is expected to be diverse.
We will now proceed by describing in detail the tasks that we defined in our benchmark.

\subsection{Associative Prediction}\label{sup:associative_pred}
We defined a total of 15 tasks involving in-context associations between two \texttt{Symbol} instances. SymbolicAI's overloaded operators rely on predefined pseudo-grammars, as described in Section \ref{formal_languages}, that augment the operators with few-shot examples. For instance, the overloaded operator \texttt{+} utilized between two \texttt{Symbol} instances provides few-shot examples how to resolve additions with various data types: 
\begin{lstlisting}[language={}]
"'1' + 2 =>3",
"17 + 'pi' =>20.1415926535...",
"7.2 + 'five' =>12.2",
"True + 0 => False",
"False + 'True' =>False",
"['a', 'b'] + ['c', 'd'] =>['a', 'b', 'c', 'd']",
"['apple'] + 'banana' =>['apple', 'banana']",
"'Zero' + 1 =>1",
"'One' + 'Two' =>3",
"'Three' + 4 =>7",
"'a + b' + 'c + d' =>a + b + c + d",
...
\end{lstlisting}
Consequently, we can now test if the models can solve the addition between \texttt{Symbol("two hundred and thirty four")} and \texttt{Symbol(7000)}.

\subsection{Multi-modal Binding}\label{sup:multimodal_binding}
We perform transformations between multiple modalities through language-based representations.
Therefore, we need to evaluate the model's proficiency in tool utilization, classification and routing of requests to relevant modules. 
We define a multi-modal \ttt{Expression} to detect the category of a task based on its content and to forward the task to the appropriate tool. The expression creates interfaces to tools like WolframAlpha for mathematical expressions, Selenium for website content scraping, SerpApi for search queries, and APILayer for optical character recognition.
Each of the five tests aims to evaluate the appropriate handling of a specific type of input by the multi-modal \ttt{Expression} type, such as processing a website URL for scraping, interpreting a search engine query, testing if two vectors are linearly independent, comparing large numbers, and extracting text from an image.
The following example shows the \ttt{MultiModalExpression} implementation of the \ttt{forward} function that uses \ttt{isinstanceof} operator on its own context to determine its current expression value and select the sub-routine that can evaluate the request.

\begin{lstlisting}[language=Python]
class MultiModalExpression(Expression):
  def forward(self, ...):
    formula = self.extract('mathematical formula')
    ...
    if self.isinstanceof(LINEAR_ALGEBRA):
      ...
      res = self.solver(formula)
      res = res.query('write a one sentence summary of the answer')
      ...
    elif self.isinstanceof(NUMBER_COMPARISON):
      res = self.solver(formula) # send directly to wolframalpha
    else:
      ...

query = Symbol("is 100044347 bigger than 129981063.472?")
expr = MultiModalExpression(query)
res = expr(...)
\end{lstlisting}

\subsection{Program Synthesis}\label{sup:program_synthesis}
We designed three separate tests related to program synthesis, where each task assesses the ability of the models to generate and execute code based on natural language instructions or provided templates:

1) The first task involves reading a LaTeX table template and data, then generating a function to populate the table with the given data. 

2) The second task tests the automatic code generation for API calls by fetching data from a specified URL and extracting specific information from the retrieved content. 

3) The third task evaluates the ability to construct a custom \ttt{Expression} that processes a \ttt{Symbol} through a specific \ttt{Function} component from the SymbolicAI package. 

\begin{figure}[!h]
    \centering
    \includegraphics[scale=0.3]{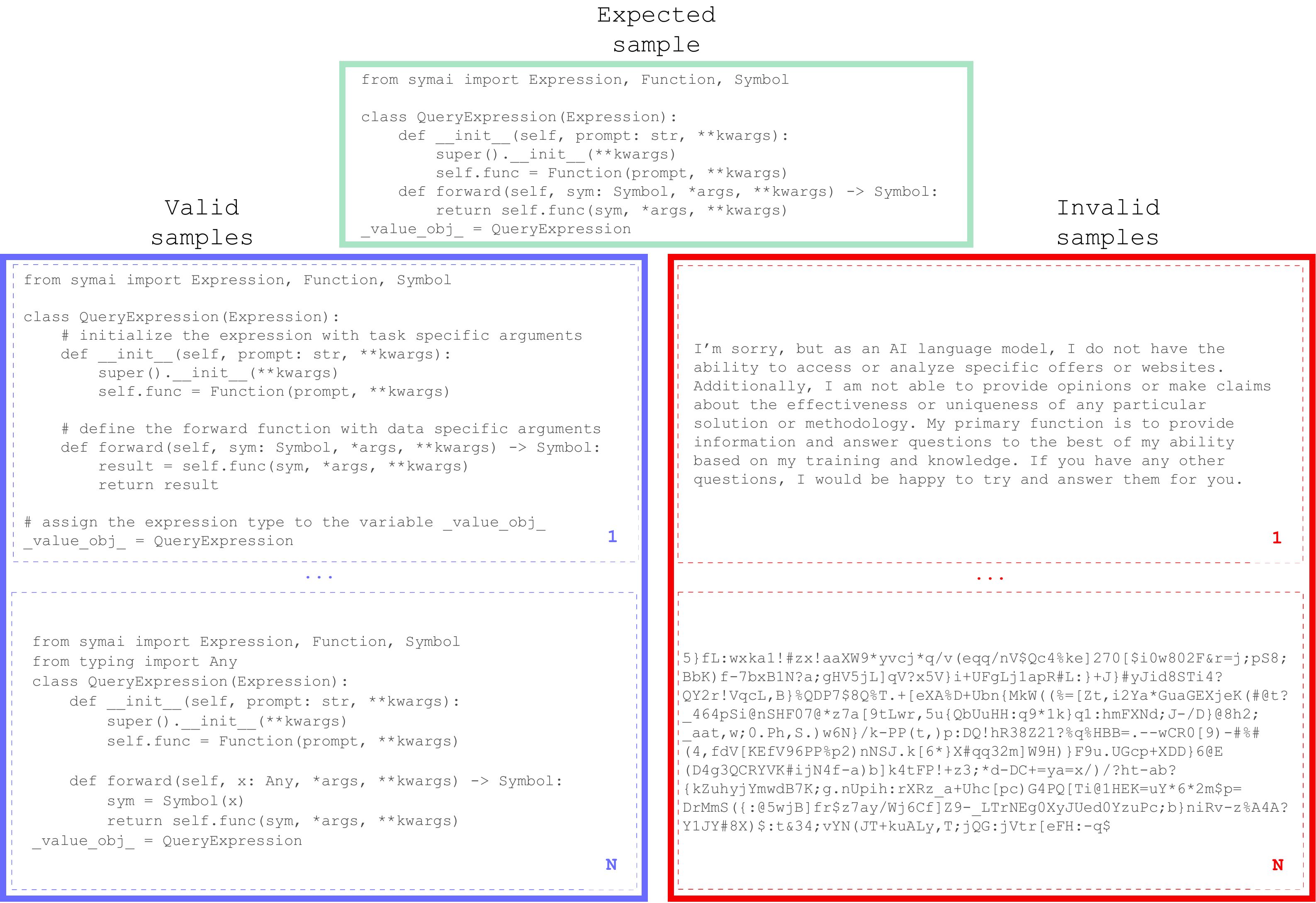}
    \caption{Samples for constructing custom expressions. The expected references are human-generated and are distributed according to the solution space. The set of valid references can be as well human-generated. The set of invalid (or undesirable) samples represent the subset we are the least interested in. Lastly, note that in general, the references need not be human-generated; any combination of human- and machine-generated references is possible. For instance, if synthetic data is utilized to distill knowledge from a larger model to a smaller model, the valid references and the expected references can be sampled from the larger model.}
    \label{fig:trajectories}
\end{figure}

Each of the three tests follows a similar pattern, where the generated code is scored based on its similarity to valid references and normalized with random samples.
Figure \ref{fig:trajectories} shows possible samples from the third task category. 

\subsection{Logical Components}\label{sup:logical_comps}
We designed six tests to assess the logical capabilities of the candidate models and group them as follows.

1) We utilize the Python library SymPy for symbolic mathematics to create the mathematical expression $ax + bx - cx - ay - by + cy + d$. The task for the model is then to factorize the expression and extract all unique symbols as a list.

\begin{lstlisting}[language=Python]
import sympy as sym

...
a, b, c, d, x, y = sym.symbols('a, b, c, d, x, y')
expr = a * x + b * x - c * x - a * y - b * y + c * y + d
# validate with sympy
fact = sym.collect(expr, d, func=sym.factor)
# model based factorization
func = Factorization('Factorize d from the expression such that your final start with: `d + (...`:')
res = func(expr)
# compare res with fact
...
\end{lstlisting}

2) Three tasks evaluate a models' capability to resolve the logical operations AND, OR, and XOR. 
For instance, the test for logical AND combines the symbols \ttt{Symbol("The horn only sounds on Sundays")} and \ttt{Symbol("I hear the horn")} and compares the answer against the human-generated references "\textit{The horn only sounds on Sundays and I hear the horn.}" and "\textit{Since I hear the horn it is Sunday.}" 
Since there is a large number of possible solutions, there is high variability in the solution space. Each model might prefer a different solution. 

The following snippet shows how one can define a custom primitive class (\ttt{CustomLogicPrimitive}) for logic operators. The \ttt{\_\_or\_\_} function gets overloaded and uses the built-in \ttt{logic} decorator from the \ttt{core} package to create a local function that evaluates two \ttt{Symbol} instances.

\begin{lstlisting}[language=Python]
from symai import core
from symai.ops.primitives import Primitive

class CustomLogicPrimitive(Primitive):
  def __or__(self, other: Any) -> Any:
    @core.logic(operator='or')
    def _func(_, a: str, b: str):
      pass # could impl. a fallback behavior here
    return self._to_symbol(_func(self, other))
  ...

subject = 'cat'
res = (Symbol(f'The {subject} has whiskers.', primitives=CustomLogicPrimitive) | \
       Symbol(f'The {subject} has a tail.', primitives=CustomLogicPrimitive)) 
\end{lstlisting}

3) For another task we use a custom \ttt{Expression} that defines a DSL syntax and semantic structure.
We use this Expression to extract higher-order logic expressions from a natural language statement, namely the puzzle 'Who is Jay's brother?'\footnote{\ Bob has two sons, John and Jay. Jay has one brother and father. The father has two sons. Jay's brother has a brother and a father. Who is Jay's brother?}, that preserves the original relationships.  The following is a DSL snippet of the 'Who is Jay's brother?' puzzle:

\begin{lstlisting}[language={}]
// Query
IsBrotherOf(jay, john, bob) <- BrotherOf(jay, john) AND FatherOf(bob, jay) AND FatherOf(bob, john);

// Facts
BrotherOf(x, y) <- HAS(x, brother) AND HAS(y, brother) AND Sibling(x, y);
FatherOf(x, y) <- HAS(x, son) AND ParentOf(x, y);
ParentOf(x, y) <- IS(x, parent) AND IS(y, child);
Sibling(x, y) <- IS(x, father) AND IS(y, father) OR IS(x, mother) AND IS(y, mother);

...
\end{lstlisting}

4) For the final task, we again use the puzzle 'Who is Jay's brother?' to evaluate a models' capability for complex conversions. 
We use the Z3 theorem prover \citep{Moura:08} to solve the 'Who is Jay's brother' puzzle conditioned on the Z3 solvers' solution to Einsteins' famous puzzle 'Who owns the fish?'.
The task involves an indirect translation from natural language to executable code by the Z3 solver; the solution to Einstein's puzzle acts as a form of self-contained "documentation" for how the Z3 solver should be utilized.
The test constructs a template, which includes the task instructions, puzzle statement, and reference to the Einstein's puzzle solution. 
The models are then asked to analyze the given problem and solution format and create a Python function with Z3 syntax that can solve the 'Who is Jay's brother?' puzzle.
The dynamically generated code is executed within the test environment utilizing Python's \ttt{exec} function. 
We check the access to the Z3 solver and run the generated \ttt{solve\_puzzle} function supposed to contain the logic to solve the puzzle.
Once executed, the assembled Z3 logical clauses are processed by the solver, which verifies that the set of constraints is satisfiable. 
If so, the model generated by the solver is queried for the puzzle's solution and scored using our VERTEX score.
The following is an example output from the Z3 representation of the solution to 'Who is Jay's brother?' puzzle:

\begin{lstlisting}[language=Python]
from z3 import Solver, Bool, And, Not, Const, BoolSort, EnumSort, Function, IntSort

def solve_puzzle(S: Solver) -> Const:
  # Define the enumeration sort for the individuals
  Person, (BobE, JohnE, JayE, JaysBrotherE, FatherE) = EnumSort('Person', ['Bob', 'John', 'Jay', 'JaysBrother', 'Father'])

  # Define a function from boolean to persons (for brother status)
  is_brother = Function('is_brother', Person, BoolSort())

  # Define the relationships
  S.add(is_brother(JohnE) == True)  # John is a brother
  S.add(is_brother(JayE) == True)   # Jay is a brother

  ...

  return query
\end{lstlisting}

\subsection{Hierarchical Computational Graphs}\label{sup:hcg}
In this section we extend on the hierarchical computational graphs section.

\paragraph{Research Paper Draft}
The following example defines a \texttt{Paper} expression that takes in a sequence of expressions which are sequentially executed. 
The \texttt{Method} expression contains a \texttt{Source} expression which points to the actual human-written method.
The \ttt{Method} expression acts as the root node that bootstraps the generation process.
The \texttt{RelatedWork} expression contains a sequence of \texttt{Cite} expressions which are executed in parallel and utilized to define the context of the related work section. 
The \ttt{Abstract} and \ttt{Title} expressions get executed last because they require all the previous information to be available in their respective contexts.
Each expression in the sequence of expressions from \texttt{Paper} takes in the context of its predecessors.
All expressions also link their results to a global linker object which is utilized after the execution to retrieve individual results from the nodes of the expression's computational graph.
Each node was evaluated against its corresponding reference, all references representing actual sections from this research paper.
The samples for each node were generated with a separate model (Claude 2) that was not part of this evaluation.
In Figure~\ref{fig:graph_result} we show the resulting computational graph of the \texttt{Paper} expression.

\begin{lstlisting}[language=Python]
# define the computational graph
expression = Paper(
  Method(
    # link to original code base where the main method is defined
    Source(file_link='/path/to/.../file'),
  ),
  # gather resources and write the related work
  RelatedWork(
    Cite(file_link='/path/to/.../file'),
    Cite(file_link='/path/to/.../file'),
    ...
  ),
  # write the abstract and title
  Abstract(),
  Title(),
)
# run the graph
paper_result = expression('Write a scientific paper')
# access linker to retreive the results from the method expression
method = expr.linker.find('Method') 
# print result of the method expression
print(method)
\end{lstlisting}

\subsection{Caveats}\label{sup:caveats}
One may have noticed that there are cases in which there is no need to sample multiple samples as there is only one expected answer, for instance, if we need to extract a specific number from a string and cast it to an integer. 
If the extraction process appends additional characters other than the number, the casting will fail.
In such cases, the VERTEX score simply defaults to the chosen similarity measure, registering and penalizing any deviation from the expected answer.
\end{document}